\documentclass[10pt,twocolumn,letterpaper]{article}

\usepackage{wacv}              

\usepackage{graphicx}
\usepackage{amsmath}
\usepackage{amssymb}
\usepackage{booktabs}

\usepackage{eccvabbrv}
\usepackage{amsmath}
\usepackage{amssymb}
\usepackage{mathtools}
\usepackage{amsthm}
\usepackage{multirow}
\usepackage{colortbl}
\usepackage{adjustbox}

\usepackage{graphicx}
\usepackage[ruled]{algorithm2e}
\usepackage{upgreek}
\usepackage{wrapfig}
\usepackage[dvipsnames]{xcolor}
\usepackage{bbding}
\usepackage{soul}
\usepackage{pifont}
\usepackage[accsupp]{axessibility}

\usepackage{graphicx}

\theoremstyle{definition}

\usepackage[ruled]{algorithm2e}
\definecolor{cvprblue}{rgb}{0.21,0.49,0.74}
\definecolor{greenx}{RGB}{0,128,128}
\definecolor{maroonx}{RGB}{195,18,48}
\usepackage[colorlinks=True,
            linkcolor=maroonx,
            anchorcolor=blue,  
            pagebackref,
            citecolor=cvprblue,
            ]{hyperref}
\usepackage{enumitem}
\setlist{leftmargin=5.5mm}
\usepackage{wrapfig}

\newcommand{\cc}{\cellcolor{gray!20}}

\setlist[itemize]{leftmargin=*,topsep=0em,itemsep=0pt}

\usepackage[capitalize]{cleveref}
\crefname{section}{Sec.}{Secs.}
\Crefname{section}{Section}{Sections}
\Crefname{table}{Table}{Tables}
\crefname{table}{Tab.}{Tabs.}

\def\wacvPaperID{1925} 
\def\confName{WACV}
\def\confYear{2025}

\begin{document}

\title{Enhancing Vision-Language Few-Shot Adaptation with Negative Learning}

\author{
Ce Zhang \quad Simon Stepputtis \quad Katia Sycara\quad Yaqi Xie\vspace{1mm}\\
School of Computer Science, Carnegie Mellon University\\ 
{\tt\small \{cezhang, sstepput, katia, yaqix\}@cs.cmu.edu} 
}

\maketitle

\begin{abstract}
Large-scale pre-trained Vision-Language Models (VLMs) have exhibited impressive zero-shot performance and transferability, allowing them to adapt to downstream tasks in a data-efficient manner. However, when only a few labeled samples are available, adapting VLMs to distinguish subtle differences between similar classes in specific downstream tasks remains challenging.
In this work, we propose a \textbf{Sim}ple yet effective \textbf{N}egative \textbf{L}earning approach, SimNL, to more efficiently exploit the task-specific knowledge from few-shot labeled samples. Unlike previous methods that focus on identifying a set of representative positive features defining ``\textit{what is a} \{\texttt{CLASS}\}", SimNL discovers a complementary set of negative features that define ``\textit{what is not a} \{\texttt{CLASS}\}", providing additional insights that supplement the positive features to enhance task-specific recognition capability.
Further, we identify that current adaptation approaches are particularly vulnerable to potential noise in the few-shot sample set. To mitigate this issue, we introduce a plug-and-play few-shot instance reweighting technique to suppress noisy outliers and amplify clean samples for more stable adaptation.
Our extensive experimental results across 15 datasets validate that the proposed SimNL outperforms existing state-of-the-art methods on both few-shot learning and domain generalization tasks while achieving competitive computational efficiency. 
Code is available at \href{https://github.com/zhangce01/SimNL}{https://github.com/zhangce01/SimNL}.
\end{abstract}

\section{Introduction}
\vspace{-2pt}
\label{sec:intro}
\noindent
Over the past decade, deep learning models have achieved remarkable progress in various vision tasks~\cite{redmon2016you,russakovsky2015imagenet}, largely due to training on extensive supervised datasets~\cite{deng2009imagenet,lin2014microsoft}. 
However, in real-world scenarios, acquiring such large-scale datasets in specific domains (\eg, satellite and aircraft images) is often impractical due to the time-consuming and costly nature of the collection and annotation process. As a result, there is often not enough data in these domains  to capture the variability of the novel classes involved, making it challenging to develop robust models from scratch that can effectively generalize to unseen data~\cite{wang2020generalizing,song2023comprehensive}.

Recent advances in Vision-Language Models (VLMs), such as CLIP~\cite{radford2021learning} and ALIGN~\cite{jia2021scaling}, provide a promising alternative approach for building robust models in a low-data regime. Specifically, by utilizing large-scale pre-training on web-scale datasets, these VLMs have demonstrated remarkable zero-shot performance and transferability~\cite{radford2021learning,yu2022coca,zhai2022lit,li2022supervision}, which enables the adaptation of these models to downstream tasks in a data-efficient manner.  Currently, researchers have developed two primary few-shot adaptation strategies for VLMs: prompt-based learning~\cite{zhou2022learning,zhou2022conditional,shu2022test,tian2023argue,zhang2024concept} and adapter-style fine-tuning~\cite{gao2021clip,zhang2022tip,li2023graphadapter,yu2023task,zhang2024fewshot}, to enable effective task-specific knowledge extraction. However, due to the very limited number of annotated samples available, these methods still struggle to discern subtle differences between similar classes in specific downstream tasks.

\begin{figure*}[t]
\centering
\includegraphics[width=\linewidth]{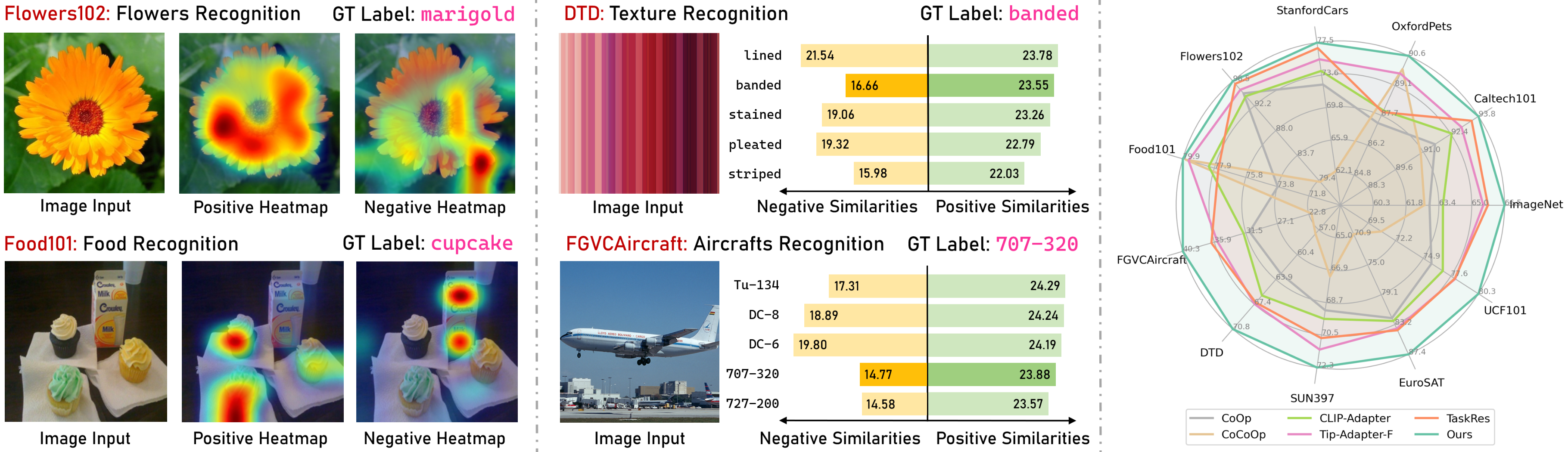}
\vspace{-17pt}
\caption{\looseness=-1 \textbf{Negative learning provides complementary information for more accurate recognition}. (\textit{Left}) Grad-CAM~\cite{selvaraju2017grad} visualization of the similarity heatmaps with the learned positive and negative features of the ground-truth class; (\textit{Middle}) Similarities (scaled by 100) to the learned positive and negative features of five similar candidate classes. While the positive branch alone may fail to distinguish among some closely related classes, incorporating the negative classifier, which eliminates certain incorrect classes, enhances the model's ability to accurately identify the true class; (\textit{Right}) Performance comparisons with other state-of-the-art methods in 16-shot scenarios.}
\label{fig:qualitative}
\vspace{-10pt} 
\end{figure*}

In this work, inspired by the negative learning literature~\cite{ishida2017learning,yu2018learning,sun2022dualcoop}, we propose a simple yet effective negative learning approach,  SimNL, for more effectively adapting VLMs to downstream tasks.
Specifically, while previous adaptation methods~\cite{zhou2022learning,zhou2022conditional,zhang2022tip} typically focus on identifying a set of representative features that define ``\textit{what is a} \{\texttt{CLASS}\}" during few-shot adaptation, we propose to discover a complementary set of negative features that define ``\textit{what is not a} \{\texttt{CLASS}\}" to better exploit the limited task-specific knowledge from the few-shot samples.
As shown in Figure~\ref{fig:qualitative} (\textit{Left}), this complementary set of negative features guides our model to pay attention to more diverse attributes when classifying an image, evaluating both  the presence of class-specific characteristics and the absence of characteristics not associated with the class.
Building on these discovered features, our SimNL framework incorporates two coordinated classifiers: one performing similarity matching with the positive features and the other performing dissimilarity matching with the negative features.
In Figure~\ref{fig:qualitative} (\textit{Middle}), we illustrate that when the positive classifier struggles with distinguishing between similar classes, employing the negative one can provide further insights to improve recognition accuracy. The performance comparisons in Figure~\ref{fig:qualitative} (\textit{Right}) further highlight that our SimNL method, by incorporating an additional negative classifier, consistently achieves superior few-shot adaptation performance across 11 various datasets when compared to existing state-of-the-art approaches.

Another important challenge often overlooked in existing adaptation methods is the presence of noise in the few-shot sample set. Most existing approaches assume that each few-shot sample is carefully curated to accurately represent its class, but such assumptions rarely holds in practice. Given the extremely small size of the few-shot sample set, the models trained from these samples are particularly vulnerable to noise. 
To address this important issue, we extend SimNL by reweighting each few-shot instance with a confidence score, which suppresses outliers and amplifies clean samples to ensure stable adaptation. This plug-and-play technique can also be applied to other adapter-style fine-tuning methods, such as Tip-Adapter-F~\cite{zhang2022tip}, making them more robust to noisy data.

To empirically validate the effectiveness of our SimNL, we conduct comprehensive evaluations on few-shot learning and domain generalization tasks across 15 diverse recognition datasets. These experimental results demonstrate the effectiveness of SimNL in adapting VLMs for downstream tasks and verify its superior robustness to distribution shifts. We also demonstrate that our proposed instance reweighting significantly enhances the model's robustness to label noises in the few-shot sample set. Additionally, SimNL also exhibits competitive efficiency.

Our primary contributions are summarized as follows: 
\begin{itemize}
    \item We propose a simple yet effective negative learning approach, \ie, SimNL, to efficiently adapt CLIP to downstream tasks. Specifically, SimNL introduces an innovative application of negative learning to adapter-style fine-tuning for few-shot adaptation of VLMs.
    \item To mitigate the impact of noisy samples in few-shot adaptation, we introduce a plug-and-play few-shot instance reweighting technique that assigns non-uniform confidence scores to each sample.
    \item  Through extensive experiments, we demonstrate that our SimNL consistently outperforms other state-of-the-art methods across 15 diverse recognition datasets.
\end{itemize}


\section{Related Work} 
\noindent
\textbf{Efficient Adaptation for VLMs}. In recent years, extensive Vision-Language Models (VLMs) have been developed to bridge the vision and language modalities through large-scale pre-training~\cite{radford2021learning,jia2021scaling}. Given the substantial size of the VLMs, recent research efforts~\cite{zhou2022learning,gao2021clip,zhang2022tip,zhang2024dual} are focusing on the development of lightweight fine-tuning techniques to efficiently adapt VLMs for downstream visual tasks. These methods can generally be divided into two categories: \textit{prompt-based learning} and \textit{adapter-style fine-tuning}. Specifically, prompt-based learning methods~\cite{zhou2022learning,zhou2022conditional,chen2023plot,tian2023argue,wu2023prompt} aim to optimize the input prompts from downstream data, while adapter-style fine-tuning approaches~\cite{gao2021clip,zhang2022tip,udandarao2023sus,yu2023task,zhu2023not} directly tune the extracted visual and textual representations. 
In this work, we aim to enhance adapter-style fine-tuning by incorporating negative learning into vision-language few-shot adaptation. We also empirically validate that leveraging negative cues from CLIP can effectively improve both few-shot classification performance and generalization capability.

\looseness=-1
\textbf{Few-Shot Learning}. Few-shot learning aims to quickly adapt a model to new categories using only a few examples. 
Traditional few-shot learning methods primarily rely on meta-learning and can be roughly separated into two groups: metric-based methods~\cite{vinyals2016matching,snell2017prototypical,zhang2020deepemd} and optimization-based methods~\cite{ravi2017optimization,rusu2018metalearning,grant2018recasting}.
However, these methods depend on training with base datasets, limiting their applicability in real-world scenarios. Recent developments in large-scale pre-trained VLMs~\cite{radford2021learning,alayrac2022flamingo} present a promising alternative due to their exceptional zero-shot capabilities. 
Researchers have demonstrated that with efficient adaptation, VLMs can also excel in few-shot learning tasks~\cite{zhou2022learning,zhou2022conditional,zhang2022tip,hu2024learning}. 

\textbf{Negative Learning}. Given that obtaining typical positive labels (which indicate the categories an image belongs to) can be costly and labor-intensive, researchers have proposed an alternative approach known as the indirect negative learning~\cite{ishida2017learning,yu2018learning} paradigm. This method focuses on learning from negative/complementary labels, which specify the categories to which an image does \textit{not} belong.
In recent years, negative learning has been effectively applied to various vision applications, such as image recognition~\cite{ishida2017learning,kim2019nlnl,gao2021discriminative}, few-shot learning~\cite{huang2022task,wei2022embarrassingly}, and semantic segmentation~\cite{wang2022semi,qiao2023fuzzy}. 
In this work, empowered by the strong negative understanding capabilities of the large-scale pre-trained CLIP model~\cite{wang2023clipn,nie2024outofdistribution,huang2024noise}, we enable an alternative approach to negative learning, where we explicitly train a negative classifier without the need for complementary labels. We provide more discussions on the differences between our method and traditional negative learning in Section~\ref{subsec:difference2}. We also note that recent work on TDA~\cite{karmanov2024efficient} shares a similar approach to ours, as both methods incorporate negative caches to enhance the generalizability of VLMs. However, their method focuses on test-time adaptation, where training data is unavailable, making it unsuitable for direct application in few-shot adaptation scenarios.

\section{Method}
\label{sec:method}
\begin{figure}[t]
\centering
\includegraphics[width=\linewidth]{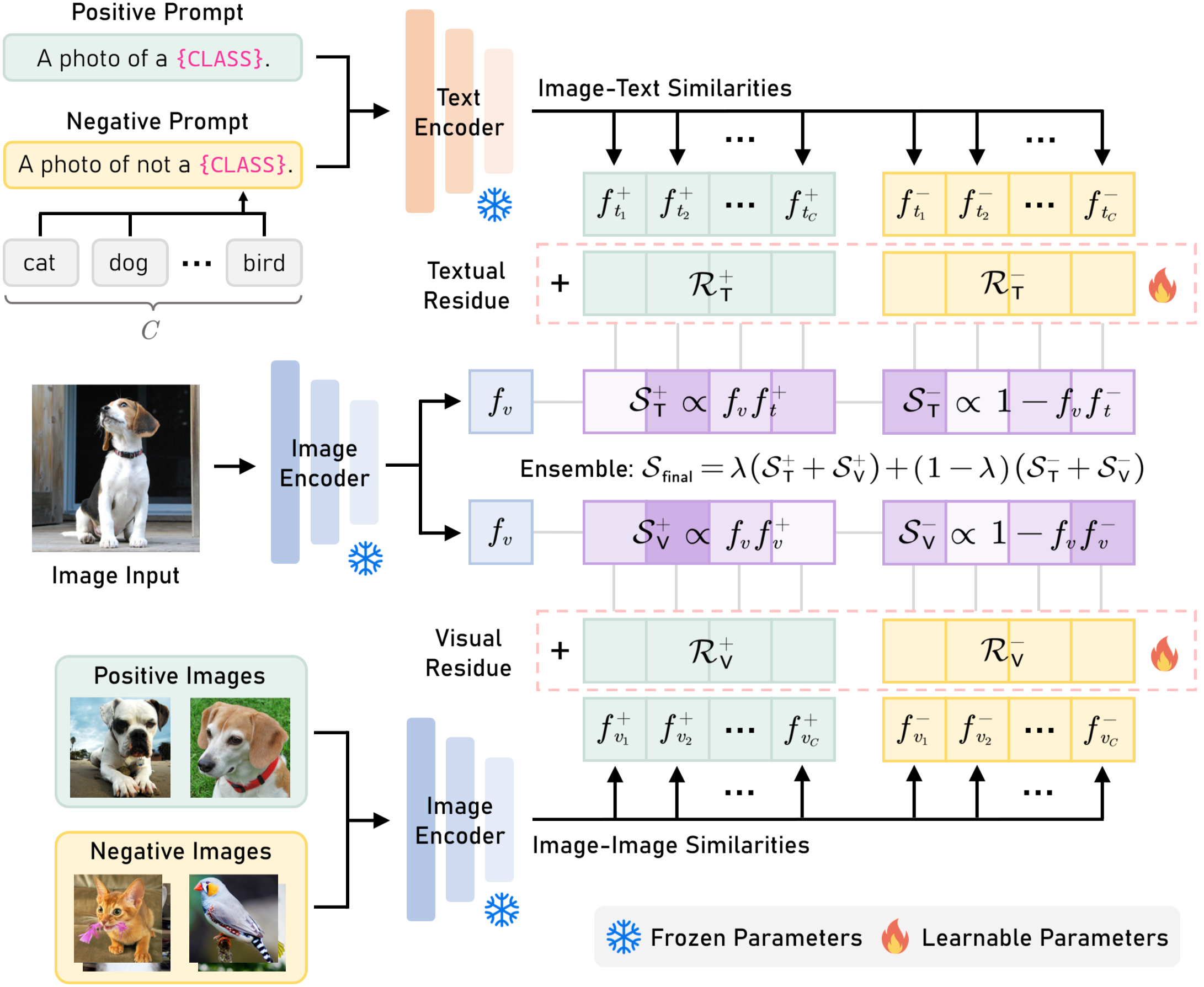}
\vspace{-15pt}
\caption{\textbf{An overview of our proposed SimNL}. We construct and learn the positive and negative CLIP-based classifiers across visual and textual modalities. Given an image to be classified, the classification logit for a specific class increases when the image feature $f_v$ closely aligns with the corresponding positive features $f_t^+, f_v^+$ and diverges from negative features $f_t^-, f_v^-$.}
\label{fig:overview}
\vspace{-5pt}
\end{figure}

\noindent
\looseness=-1
In this section, we introduce SimNL, a novel approach to vision-language few-shot adaptation, as shown in Figure~\ref{fig:overview}.

\subsection{Problem Formulation}
\label{sec:formulation}
\noindent
We consider the problem of $C$-way $K$-shot few-shot classification problem, in which only $K$ labeled examples are provided for each of the $C$ classes. Let the feature space be  $\mathcal{X}$ and the label space be $\mathcal{Y} = \{ 1, ..., C\}$, the few-shot instance  $\boldsymbol{x} \in \mathcal{X}$ and its labels $y \in \mathcal{Y}$ are sampled from probability distribution $P(\boldsymbol{x}, y)$. The corresponding one-hot encoded labels are represented by $\boldsymbol{y} \in \{0, 1\}^C$. 

\looseness=-1
\textbf{Positive Learning}. In this study, we employ a CLIP-based classifier with parameters $\theta$, denoted as $\mathcal{F}_{\theta}: \mathcal{X} \rightarrow \mathbb{R}^C$, to project each input into a $C$-dimensional score space. The classification probabilities are obtained using $\boldsymbol{p} = \texttt{Softmax}\left(\mathcal{F}_{\theta}(\boldsymbol{x})\right) \in \Delta^{C-1}$, where $\Delta^{C-1}$ represents the $(C-1)$-dimensional probability simplex. Our goal is to learn a robust classifier $\mathcal{F}_{\theta}$ that minimizes the expected risk; that is,
\begin{gather}
\setlength\abovedisplayskip{3pt}
\setlength\belowdisplayskip{3pt}
    \min_{\theta} \mathcal{R}(\mathcal{F}_{\theta})= \mathbb{E}_{(\boldsymbol{x}, y) \sim P(\boldsymbol{x}, y)} \left[\mathcal{L}\left(\mathcal{F}_{\theta}(\boldsymbol{x}), y \right)\right], \nonumber\\ \text{where}\,\, \mathcal{L}\left(\mathcal{F}_{\theta}(\boldsymbol{x}), y \right) = - \sum_{k=1}^{C} \boldsymbol{y}_k \log \boldsymbol{p}_k,
    \label{eq:positive}
\end{gather}
where $\boldsymbol{p}_k$ denotes the probability for the $k$-th class and $\mathcal{L}\left(\mathcal{F}_{\theta}(\boldsymbol{x}), y \right)$ computes the cross-entropy loss.  We aim to maximize the probability $\boldsymbol{p}_y \rightarrow 1$ for the 
true label.


\textbf{Negative Learning}. In this work, we apply the concept of negative learning to vision-language few-shot adaptation by employing a distinct CLIP-based negative classifier $\mathcal{G}_{\varphi}: \mathcal{X} \rightarrow \mathbb{R}^C$ with parameters $\varphi$. This classifier predicts the negative probability $\bar{\boldsymbol{p}} = \texttt{Softmax}(\mathcal{G}_{\varphi}(\boldsymbol{x}))$ that the image does not belong to specific classes. The expected classification risk for  $\mathcal{G}_{\varphi}$ can be written as
\begin{gather}
\setlength\abovedisplayskip{3pt}
\setlength\belowdisplayskip{3pt}
    \min_{\varphi} \mathcal{R}(\mathcal{G}_{\varphi})= \mathbb{E}_{(\boldsymbol{x}, y) \sim P(\boldsymbol{x}, y)} \left[\mathcal{L}\left(\mathcal{G}_{\varphi}(\boldsymbol{x}), y \right)\right],\nonumber \\ \text{where}\,\, \mathcal{L}\left(\mathcal{G}_{\varphi}(\boldsymbol{x}), y \right) = - \sum_{k=1}^{C} \boldsymbol{y}_k \log (1 - \bar{\boldsymbol{p}}_k).
\end{gather}
By optimizing this risk, we aim to reduce the negative probability $\bar{\boldsymbol{p}}_{y} \rightarrow 0$ for the true label. 

Finally, we ensemble the optimized classifiers from both positive and negative learning to create an enhanced classifier $\mathcal{F}_{\theta} \oplus \mathcal{G}_{\varphi}: \mathcal{X} \rightarrow \mathbb{R}^C$ for testing. This is equivalent to combining the positive and negative predicted probabilities as $\hat{\boldsymbol{p}} = \lambda \boldsymbol{p} + (1-\lambda) (1 - \bar{\boldsymbol{p}})$, where $\lambda$ is a balancing hyper-parameter.

\looseness=-1
\textbf{Differences with Contrastive Learning}. Note that the concept of ``negative" differs in our negative learning approach compared to contrastive learning:
(1) In our negative learning, ``negative" specifically refers to a negative classifier. We explicitly train another negative classifier $\mathcal{G}_{\varphi}$ to ensemble with the positive classifier; (2) In contrastive learning, ``negative" refers to the negative sample pairs that are constructed and utilized during training. In Appendix~\ref{subsec:difference}, we show that the training objectives for both the positive and negative CLIP-based classifiers are contrastive in nature.




\subsection{Preliminary: A Revisit of CLIP}
\label{sec:clip}
\noindent
In this work, our classifiers are based on CLIP's pre-trained visual encoder $\mathcal{F}_\mathsf{V}$ and textual encoder $\mathcal{F}_\mathsf{T}$. Specifically, CLIP~\cite{radford2021learning} performs zero-shot predictions by assessing the similarity between the image feature and the text feature specific to each class $c\in \{ 1, ..., C\}$  as follows:
\begin{gather}
\setlength\abovedisplayskip{3pt}
\setlength\belowdisplayskip{3pt}
\label{eq:clip}
f_{v}=\mathcal{F}_\mathsf{V}(\boldsymbol{x}), \quad f_{t_c}=\mathcal{F}_\mathsf{T}(\mathcal{T}_c), \nonumber\\
\mathbb{P}(y=c|\boldsymbol{x}) = \frac{\exp \left( \mathrm{cos}\left(f_{t_c},f_{v} \right) /t \right)}{\sum\nolimits_{t'}{\exp \left( \mathrm{cos}\left( f_{t'},f_{v} \right) /t \right)}}, 
\end{gather}
where $\boldsymbol{x}$ is the input image, and $\mathcal{T}_c$ represents the text description for class $c$ (\eg, “\texttt{A photo of a} \{\texttt{CLASS$_c$}\}”). The parameter $t$ is the temperature hyperparameter, and $\mathrm{cos}\left( f_{t},f_{v} \right) = f_{v}^\top f_{t}$ computes the cosine similarity. 
To streamline this process, we can precompute a textual cache, which concatenates the textual features associated with each class, denoted as $\mathsf{T}_{\mathtt{cache}} = [f_{t_1}\, f_{t_2}\, \cdots \, f_{t_C}]^\top \in \mathbb{R}^{C \times d}$. Subsequently, we can obtain the final prediction $\mathbb{P}(y|\boldsymbol{x})$ via vectorized similarity matching:
\begin{equation}
\setlength\abovedisplayskip{3pt}
\setlength\belowdisplayskip{3pt}
\label{eq:cache}
\mathcal{S} = f_{v} \mathsf{T}_{\mathtt{cache}}^{\quad\,\,\,\, \top} \in \mathbb{R}^{C}, \quad \mathbb{P}(y|\boldsymbol{x}) = \mathtt{Softmax} \left( \mathcal{S}\right).
\end{equation}

\subsection{Learning the Positive Classifier}
\noindent
\looseness=-1
In this section, we focus on the \textit{positive} perspective, namely, constructing and optimizing a learnable CLIP-based positive classifier. 
This similarity-based classifier is designed to identify a set of representative features that enhance the accuracy of predicting the true class of the input image.


\textbf{Positive Textual Branch}. As shown in Eq.~(\ref{eq:cache}), CLIP can make zero-shot predictions utilizing a textual cache, which stores the textual features from positive text descriptions (\eg, “\texttt{A photo of a\,\,\{\texttt{CLASS}\}}”). To avoid ambiguity, we denote this cache as $\mathsf{T}_{\mathtt{cache}}^{+} = [f_{t_1}^{+}\, f_{t_2}^{+}\, \cdots \, f_{t_C}^{+}]^\top \in \mathbb{R}^{C \times d}$.
With a small set of annotated training images, we can update the textual features by  introducing a group of learnable residual parameters $\mathcal{R}_{\mathsf{T}}^{+} \in \mathbb{R}^{C\times d}$ to integrate task-specific knowledge:
\begin{gather}
\mathsf{T}_{\mathtt{cache}}^{+} \leftarrow \mathtt{Normalize}\left(\mathsf{T}_{\mathtt{cache}}^{+} + \mathcal{R}_{\mathsf{T}}^{+}\right), \nonumber\\ \mathcal{S}_{\mathsf{T}}^{+} = f_v \mathsf{T}_{\mathtt{cache}}^{+\quad\top} \in \mathbb{R}^{C}. 
\end{gather}
Here, $\mathtt{Normalize}$ denotes the L2-normalization applied to each row of the matrix.

\textbf{Positive Visual Branch}. We further extend the recognition capability of the CLIP model by constructing a visual cache-based classifier, which operates by measuring image-image similarities between the input image feature and few-shot image features. Specifically, we store all $CK$ image features in a precomputed visual cache, denoted as $\mathsf{V}_{\mathtt{cache}}^{+} = [f_{v_1}^{(1)}\, f_{v_1}^{(2)}\, \cdots \, f_{v_C}^{(K)}]^\top \in \mathbb{R}^{CK \times d}$. Their one-hot labels are also correspondingly recorded in $L \in \mathbb{R}^{CK \times C}$. 
Given an image feature $f_v$ to be classified, we calculate its image-image affinities and obtain the  logits $\mathcal{S}_{\mathsf{V}}^{+}$:
\begin{gather}
\mathsf{V}_{\mathtt{cache}}^{+} \leftarrow \mathsf{Normalize}\left(\mathsf{V}_{\mathtt{cache}}^{+} + \mathcal{R}_{\mathsf{V}}^{+}\right), \nonumber\\ \mathcal{S}_{\mathsf{V}}^{+} = \mathcal{A}\left(f_v \mathsf{V}_{\mathtt{cache}}^{+ \quad \top}\right) L \in \mathbb{R}^{C},
\label{eq:positivevisual}
\end{gather}
where $\mathcal{A}\left(f_v \mathsf{V}_{\mathtt{cache}}^{+ \quad \top}\right)=\alpha \exp \left( - \beta \left( 1 - f_v \mathsf{V}_{\mathtt{cache}}^{+ \quad \top} \right)\right)$ calculates the affinity, $\alpha$ represents a balance scalar and $\beta$ denotes a modulating parameter. Learnable residue $\mathcal{R}_{\mathsf{V}}^{+} \in \mathbb{R}^{C\times d}$ is also introduced, which is broadcast to $\mathbb{R}^{CK\times d}$ and added to the visual cache to refine the visual features.


\subsection{Learning the Negative Classifier}
\noindent
Having introduced the positive classifier, we now explore constructing and learning a \textit{negative} dissimilarity-based classifier, which directs CLIP towards more confidently excluding incorrect classes based on the given input image. Conceptually, our goal is to mine a general negative feature for each class $c$, which is absent in samples from class $c$ but present in samples from all other classes. 

\textbf{Negative Textual Branch}. 
Recall that the positive textual cache, denoted as $\mathsf{T}_{\mathtt{cache}}^{+}$, is constructed based on the positive class descriptor prompt such as “\texttt{A photo of a} \{\texttt{CLASS}\}.” Conversely, we introduce a set of negative prompts (\eg, “\texttt{A photo of not a} \{\texttt{CLASS}\},” “\texttt{A photo without} \{\texttt{CLASS}\}”) that convey opposite semantics meanings.
By leveraging the textual features $f_{t}^{-}$ derived from the negative text descriptions linked with these prompts, we again precompute a weight matrix and construct negative textual cache $\mathsf{T}_{\mathtt{cache}}^{-} = [f_{t_1}^{-}\, f_{t_2}^{-}\, \cdots \, f_{t_C}^{-}]^\top \in \mathbb{R}^{C \times d}$.
Intuitively, if the feature of an input image $f_v$ closely resembles the negative textual feature for a specific class $c$, it strongly suggests that the image does not belong to class $c$, \ie, $\mathbb{P}(y=c|\boldsymbol{x}) \propto 1 - f_{v}^\top f_{t_c}^{-}$. 
Through supervised task-specific training, we optimize a learnable residue $\mathcal{R}_{\mathsf{T}}^{-} \in \mathbb{R}^{C \times d}$ and obtain the negative predictions based on dissimilarities $1 - f_{v}\mathsf{T}_{\mathtt{cache}}^{-}$:
\begin{gather}
\mathsf{T}_{\mathtt{cache}}^{-} \leftarrow \mathsf{Normalize}\left(\mathsf{T}_{\mathtt{cache}}^{-} + \mathcal{R}_{\mathsf{T}}^{-}\right), \nonumber\\ \mathcal{S}_{\mathsf{T}}^{-} = \delta_\mathsf{T} \left(1 - f_v \mathsf{T}_{\mathtt{cache}}^{- \quad \top}\right) \in \mathbb{R}^{C},
\end{gather}
where $\delta_\mathsf{T}$ is a fixed scaling parameter that adjusts $\mathcal{S}_{\mathsf{T}}^{-}$ to match the mean value of $\mathcal{S}_{\mathsf{T}}^{+}$.

\begin{figure}[t]
\centering
\includegraphics[width=\linewidth]{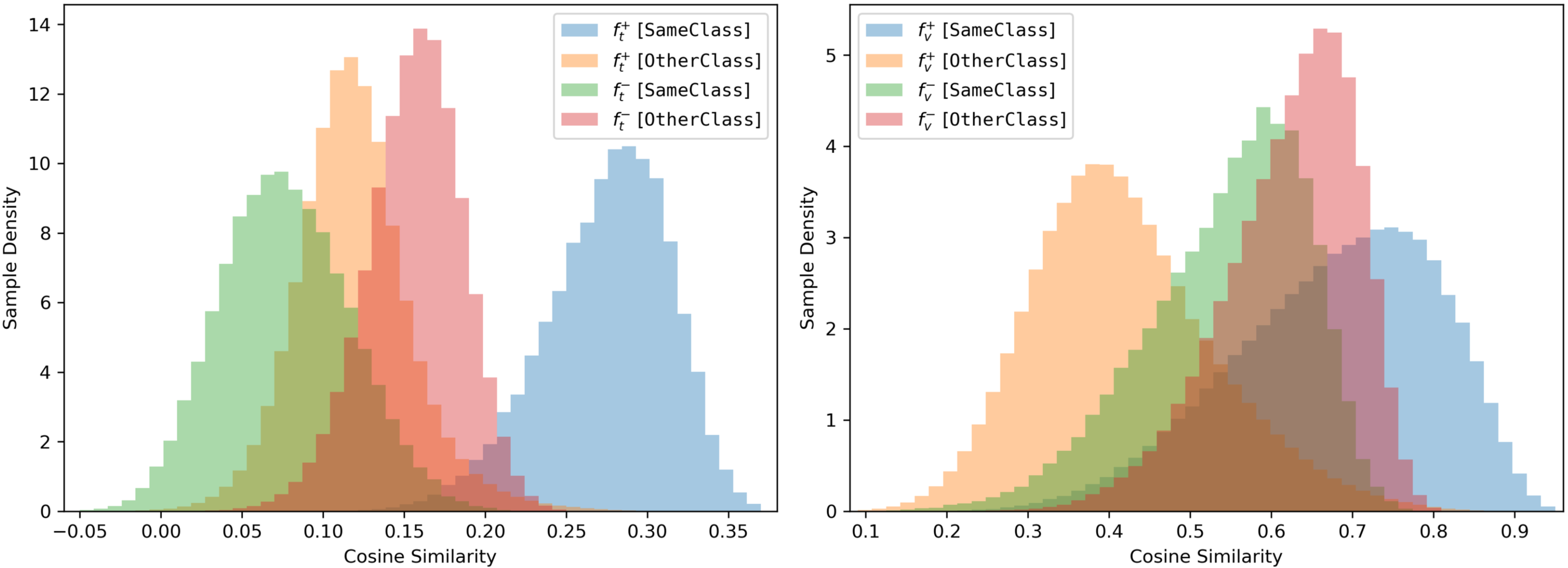}
\vspace{-15pt}
\caption{\textbf{Visualization of cosine similarities on  ImageNet~\cite{deng2009imagenet} validation set}. We present distributions of pairwise similarities between the input image feature and both the learned positive and negative features from textual (\textit{Left}) and visual (\textit{Right}) modalities.}
\label{fig:distribution}
\vspace{-8pt}
\end{figure}

\looseness=-1
\textbf{Negative Visual Branch}. In the visual domain, we also similarly pursue some negative image features to be non-representative of a specific class, which means that the higher the similarity to them, the lower the probability that the image is classified to that class, \ie, $\mathbb{P}(y=c|\boldsymbol{x}) \propto 1 - f_{v}^\top f_{v_c}^{-}$. 
To achieve this, we  randomly select one image from each of the $C-1$ classes and compute the average of their extracted features to represent the negative features. 
In this way, we can get a total of $K$ negative visual features for each of the $C$ classes, thereby constructing a negative visual cache $\mathsf{V}_{\mathtt{cache}}^{-}  \in \mathbb{R}^{CK \times d}$.
Symmetric to Eq.~(\ref{eq:positivevisual}), we compute the negative affinities to obtain the classification logits:
\begin{gather}
\mathsf{V}_{\mathtt{cache}}^{-} \leftarrow \mathsf{Normalize}\left(\mathsf{V}_{\mathtt{cache}}^{-} + \mathcal{R}_{\mathsf{V}}^{-}\right), \nonumber\\ 
\mathcal{S}_{\mathsf{V}}^{-} = \delta_\mathsf{V} \mathcal{A}\left(1 - f_v \mathsf{V}_{\mathtt{cache}}^{- \quad \top}\right) L \in \mathbb{R}^{C}.
\label{eq:negativevisual}
\end{gather}
where $\delta_\mathsf{V}$ is another fixed scaling parameter that adjusts $\mathcal{S}_{\mathsf{V}}^{-}$ to match the mean value of $\mathcal{S}_{\mathsf{V}}^{+}$. Here, we also introduce a set of learnable parameters $\mathcal{R}_{\mathsf{V}}^{-} \in \mathbb{R}^{C\times d}$ to refine the negative visual features.


\textbf{Final Inference}. 
As discussed in Section~\ref{sec:formulation}, we ensemble the predictions from both  classifiers to derive the final classification scores and predictions: 
\begin{gather}
\setlength\abovedisplayskip{3pt}
\setlength\belowdisplayskip{3pt}
\mathcal{S}_{\mathsf{final}}  = \lambda \left(\mathcal{S}_{\mathsf{T}}^{+} + \mathcal{S}_{\mathsf{V}}^{+}\right) + (1 - \lambda)\left(\mathcal{S}_{\mathsf{T}}^{-} + \mathcal{S}_{\mathsf{V}}^{-}\right), \nonumber\\ \mathbb{P}(y|\boldsymbol{x}) = \mathtt{Softmax} \left( \mathcal{S}_{\mathsf{final}}\right).
\label{eq:lambda}
\end{gather}
Here, $\lambda$ serves as a tuning hyper-parameter to balance the contribution of positive and negative logits. Throughout the training process, the collection of learnable parameters $\mathcal{R}=\{\mathcal{R}_{\mathsf{T}}^{+}, \mathcal{R}_{\mathsf{V}}^{+}, \mathcal{R}_{\mathsf{T}}^{-}, \mathcal{R}_{\mathsf{V}}^{-}\}$ is updated via stochastic gradient descent with a cross-entropy loss.


\looseness=-1
\textbf{Qualitative Visualizations}. Figure~\ref{fig:distribution} qualitatively shows the effectiveness of both the positive and negative classifiers we designed. 
Specifically, we visualize the distributions of pairwise cosine similarities between the input image feature $f_v$ and both the learned positive and negative features from two modalities ($f_t^+,f_t^-,f_v^+, f_v^-$) on the validation set of ImageNet~\cite{deng2009imagenet}. 
We can observe the following statistical patterns: (1) As we expected, the input image feature is more similar with positive features from the same class (\textcolor{Cyan}{blue} $>$ \textcolor{Peach}{orange}) and negative features from other classes (\textcolor{OrangeRed}{red} $>$ \textcolor{LimeGreen}{green}) across both two modalities;  (2) Within the visual modality, the similarity distribution of negative features occupies an intermediate position between the positive features of the same and different classes (\textcolor{Peach}{orange} $<$ \textcolor{LimeGreen}{green}/\textcolor{OrangeRed}{red} $<$ \textcolor{Cyan}{blue}). This is because the negative visual features are constructed by averaging the image features across all other classes, inherently leading to a more generic representation that lacks the distinctiveness characteristic of positive features within a single class; (3) Within the textual modality, negative features tend to be less similar to the input image compared to positive features (\textcolor{OrangeRed}{red} $<$ \textcolor{Cyan}{blue}, \textcolor{LimeGreen}{green} $<$ \textcolor{Peach}{orange}), since the negative prompts are less common in CLIP~\cite{radford2021learning} training corpus.


\subsection{Few-Shot Instance Reweighting}
\begin{figure}[t]
\centering
\includegraphics[width=\linewidth]{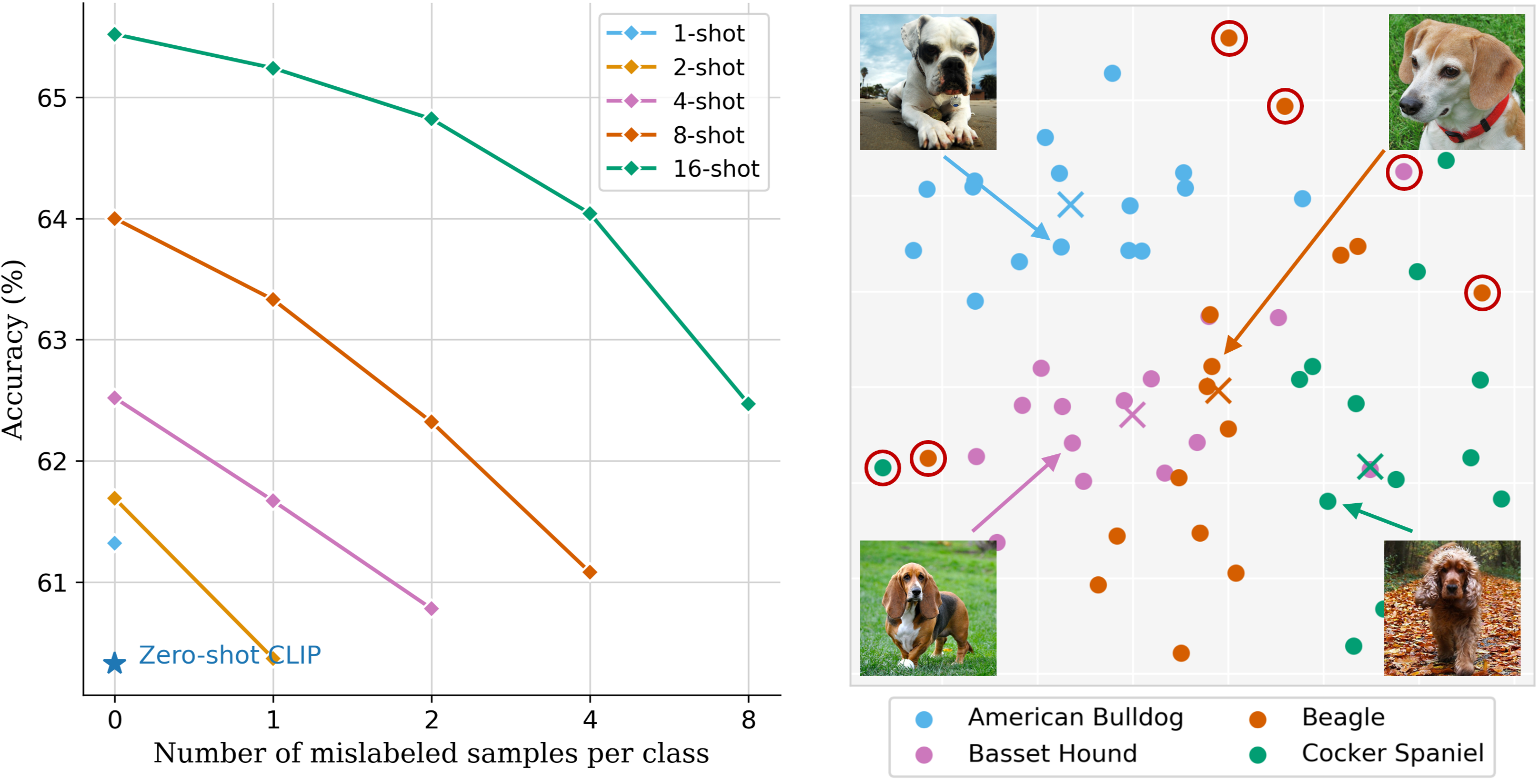}
\vspace{-15pt}
\caption{\looseness=-1 \textbf{Few-shot instance reweighting}. (\textit{Left}) The performance of Tip-Adapter-F~\cite{zhang2022tip} degrades drastically when label noise exists in the few-shot sample set; (\textit{Right}) t-SNE~\cite{van2008visualizing} visualization of visual features for 4 random classes from the OxfordPets~\cite{parkhi2012cats} dataset, where some outliers are marked with red circles.}
\label{fig:refinement}
\vspace{-8pt}
\end{figure}

\noindent
In the few-shot adaptation setting, our classifier is trained using only a limited number of samples, with each sample making a significant contribution to the formation of the final decision boundary.
Consequently, our VLMs are particularly vulnerable to potential noise in the few-shot sample set. In Figure~\ref{fig:refinement} (\textit{Left}), we simulate real-world noise scenarios by randomly flipping the labels of a portion of support samples. We demonstrate that the performance of Tip-Adapter-F~\cite{zhang2022tip} drastically decreases from 65.52\% to 62.47\% when the labels of 8 out of the 16 samples per class are randomly flipped. Moreover, even if the labels are all correct, we recognize that not every image is of high quality or equally representative of its respective class, as shown in Figure~\ref{fig:refinement} (\textit{Right}). To address this issue, we have developed a few-shot instance reweighting technique to assign non-uniform confidence scores to each sample, effectively downweighting outliers (or mislabelled samples) and prioritizing more representative samples.

Specifically, our proposed instance reweighting is based on an intuitive assumption: the representative image feature is closer to other image features from the same class than those low-quality outliers. Based on this assumption, given the $K$-shot image features $\{f_{v_c}^{(i)}\}_{i=1}^K$ from a specific class $c$, we calculate the average cosine similarities of each image feature to others:
\begin{equation}
    \mathsf{d}_c^{(i)}\!=\!\frac{1}{K-1}\!\sum_{j, \,j\neq i}\!\cos \left(f_{v_c}^{(i)},f_{v_c}^{(j)} \right)\!=\!\frac{1}{K-1}\!\sum_{j, \,j\neq i}\!f_{v_c}^{(i)\top}\!f_{v_c}^{(j)}.
\end{equation}
We then assign non-uniform weights $w_c^{(i)}$ and compute the reweighted confidences $\ell _c^{(i)}$ for all $K$-shot image features based on their average similarities to others:
\begin{equation}
\setlength\abovedisplayskip{3pt}
\setlength\belowdisplayskip{3pt}
    w_c^{(i)} = \frac{\exp(\mathsf{d}_c^{(i)} / \tau)}{\sum_{i'}\exp(\mathsf{d}_c^{(i')} / \tau)}, \quad \ell _c^{(i)} = K w_c^{(i)}, \label{eq:refine}
\end{equation}
where $\tau$ is a temperature hyper-parameter to control the intensity of our instance reweighting. 

\looseness=-1
This process is applied across each class, wherein the original one-hot labels $L$, are reweighted using the new confidence values calculated in Eq.~(\ref{eq:refine}) to yield $\mathbb{L}$. By incorporating this reweighting technique into the visual branches of both classifiers, we can reformulate Eqs.~(\ref{eq:positivevisual}) and (\ref{eq:negativevisual})  as follows:
\begin{equation}
    \mathcal{S}_{\mathsf{V}}^{+} = \mathcal{A}\left(f_v \mathsf{V}_{\mathtt{cache}}^{+ \quad \top}\right) \mathbb{L}^+  , \, \mathcal{S}_{\mathsf{V}}^{-} = \delta_\mathsf{V} \mathcal{A}\left(1 - f_v \mathsf{V}_{\mathtt{cache}}^{- \quad \top}\right) \mathbb{L}^-.
\end{equation}



\section{Experiments}
\label{sec:exp}
\noindent
In this section, we conduct extensive experiments on two tasks across 15 datasets. These results demonstrate that our proposed method is simple yet highly effective, surpassing other state-of-the-art methods in both few-shot adaptation and domain generalization capabilities.

\subsection{Experimental Settings}
\label{subsec:exp-setting}

\begin{figure*}[t]
\centering
\includegraphics[width=\linewidth]{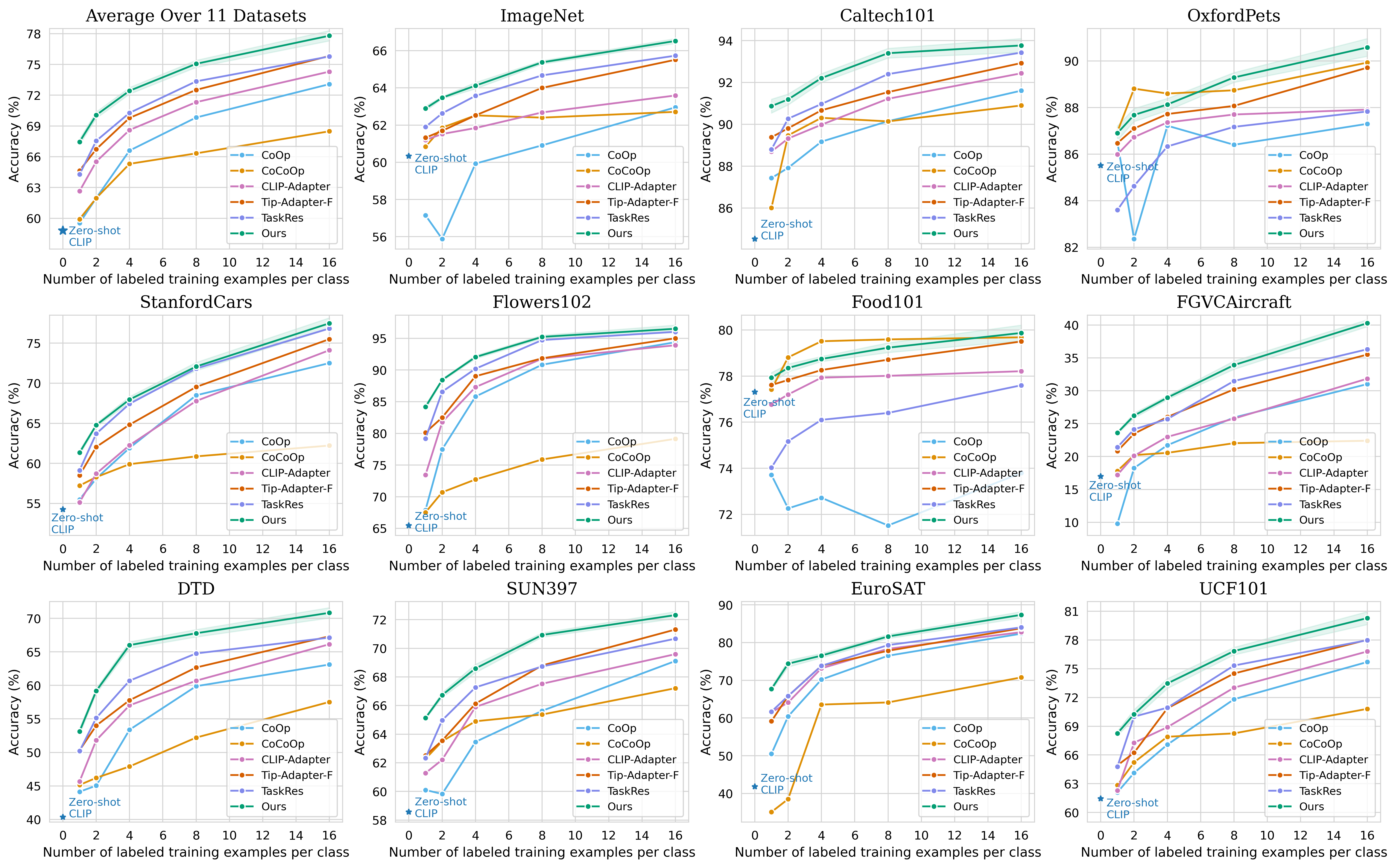}
\vspace{-20pt}
\caption{\textbf{Performance comparisons on few-shot learning on 11 image classification datasets}. For each dataset, we report the mean accuracy and 95\% confidence interval over 3 random seeds of our SimNL on 1-/2-/4-/8-/16-shot settings.}
\label{fig:fewshot}
\vspace{-12pt}
\end{figure*}

\noindent
\textbf{Tasks and Datasets}. To validate the effectiveness of SimNL, we evaluate our method on two standard benchmarking tasks: few-shot learning and domain generalization, respectively. For few-shot learning task, we comprehensively evaluate our method on 11 well-known image classification benchmarks: ImageNet~\cite{deng2009imagenet}, Caltech101~\cite{fei2004learning},  OxfordPets~\cite{parkhi2012cats}, StandfordCars~\cite{krause20133d}, Flowers102~\cite{nilsback2008automated}, Food101~\cite{bossard2014food}, FGVCAircraft~\cite{maji2013fine},  DTD~\cite{cimpoi2014describing}, SUN397~\cite{xiao2010sun}, EuroSAT~\cite{helber2019eurosat}, and UCF101~\cite{soomro2012ucf101}.
For domain generalization, we evaluate the generalizability of our SimNL on 4 variants of ImageNet: ImageNet-V2~\cite{recht2019imagenet}, ImageNet-Sketch~\cite{wang2019learning}, ImageNet-A~\cite{hendrycks2021natural}, and ImageNet-R~\cite{hendrycks2021many}. Moreover, we also explore the adaptation of VLMs using a noisy few-shot sample set from ImageNet~\cite{deng2009imagenet}, where labels are randomly flipped to simulate real-world scenarios.

\looseness=-1
\textbf{Implementation Details}. Following previous works, we adopt ResNet-50~\cite{he2016deep} backbone as the visual encoder of CLIP in our experiments by default. We adopt prompt ensembling, leveraging textual prompts from CLIP~\cite{radford2021learning}  to enhance model performance. For the negative prompts we used for each dataset, please kindly refer to Appendix~\ref{sec:prompt}. We set the hyper-parameters $\lambda$ and $\tau$ as $0.75$ and $1$, respectively. Our SimNL is trained using the AdamW~\cite{loshchilov2018decoupled} optimizer with a cosine scheduler~\cite{loshchilov2016sgdr}. The batch size is set to 256. For $\mathcal{R}_{\mathsf{T}}^{+}$ and $\mathcal{R}_{\mathsf{V}}^{+}$, the learning rate is set to $0.0001$, while for $\mathcal{R}_{\mathsf{T}}^{-}$ and $\mathcal{R}_{\mathsf{V}}^{-}$, the learning rate is set to $0.0005$. Our model is trained for 200 epochs on the EuroSAT~\cite{helber2019eurosat} dataset, and for 20 epochs on all other datasets.
To ensure the reliability of our results, we perform each experiment three times using different initialization seeds and report the mean accuracy achieved. All experiments are conducted on a single 48GB NVIDIA RTX 6000 Ada GPU. 

\textbf{Baselines}. We compare our proposed method with the following state-of-the-art methods: zero-shot and linear probe CLIP~\cite{radford2021learning}, CoOp~\cite{zhou2022learning}, CoCoOp~\cite{zhou2022conditional}, ProGrad~\cite{zhu2023prompt}, CLIP-Adapter~\cite{gao2021clip}, Tip-Adapter-F~\cite{zhang2022tip}, TPT~\cite{shu2022test}, TaskRes~\cite{yu2023task}, and GraphAdapter~\cite{li2023graphadapter}. For a fair comparison, we directly report the results of these baselines from their respective original papers.

\subsection{Results and Analysis}
\label{subsec:results}
\noindent
\textbf{Few-Shot Learning}. In Figure~\ref{fig:fewshot}, we compare the few-shot learning performance of our proposed method with other state-of-the-art methods on 11 image classification datasets. In the top-left sub-figure, we also present the average classification accuracy across all 11 datasets. The results indicate that our method consistently outperforms other methods across various few-shot learning settings by substantial margins. 
Moreover, our proposed method demonstrates more pronounced performance improvements in specialized classification tasks, such as satellite image and texture classification on the EuroSAT~\cite{helber2019eurosat} and DTD~\cite{cimpoi2014describing} datasets. With 16-shot training on these two datasets, our method surpasses Tip-Adapter-F~\cite{zhang2022tip} by a notable 3.56\% and 3.50\%, respectively. 
For full numerical results, please refer to Table~\ref{tab:numerical_results} in Appendix~\ref{sec:full}.
Overall, the consistently superior performance on 11 datasets fully demonstrates the general effectiveness of our proposed approach.

\looseness=-1
\textbf{Robustness to Natural Distribution Shifts}. In Table \ref{table:generalization}, we compare the generalizability of our SimNL with other methods in the presence of distribution shifts. Specifically, all the models are trained solely on 16-shot ImageNet~\cite{deng2009imagenet}, and directly tested on 4 out-of-distribution ImageNet variant datasets. As shown in Table \ref{table:generalization}, our method not only achieves state-of-the-art performance on the source dataset but also attains an average performance gain of 1.18\% across 4 out-of-distribution (OOD) target datasets.
These experimental results indicate that by enabling adaptation from both positive and negative perspectives, our method enhances robustness against distribution shifts.

\begin{table}
\small
\begin{center}
\caption{\looseness=-1 \textbf{Performance comparison on robustness to distribution shifts}. All the models are trained on 16-shot ImageNet~\cite{deng2009imagenet} and directed tested on the OOD target datasets. The best results are in \textbf{bold} and the second best are \underline{underlined}.}
\vspace{-5pt}
\setlength\tabcolsep{4pt}
\label{table:generalization}
\resizebox{\linewidth}{!}{
\begin{tabular}{lcccccc}
\toprule
\multirow{2}{*}[-0.5ex]{Method}  & Source & \multicolumn{5}{c}{Target} \\ \cmidrule(lr){2-2} \cmidrule(lr){3-7}  & ImageNet & -V2 & -Sketch & -A & -R  & Avg. \\
\midrule
Zero-Shot CLIP~\cite{radford2021learning}  &  60.33 &  53.27  & 35.44  & 21.65 &  56.00  & 41.59\\
Linear Probe CLIP~\cite{radford2021learning}  & 56.13  & 45.61  & 19.13  & 12.74  & 34.86 &  28.09\\
CoOp~\cite{zhou2022learning}   & 62.95  & 55.40  & 34.67  & 23.06  & 56.60 &  42.43\\
CoCoOp~\cite{zhou2022conditional} &   62.71  & 55.72  & 34.48  & 23.32  & 57.74  & 42.82\\
ProGrad~\cite{zhu2023prompt} &   62.17  & 54.70  & 34.40  & 23.05  & 56.77  & 42.23\\
TPT~\cite{shu2022test} &   60.74 &  54.70  & 35.09  & \textbf{26.67}  & 59.11  & 43.89\\
TaskRes~\cite{yu2023task} &   64.75 &  56.47  & 35.83  & 22.80  & 60.70  & 43.95\\
GraphAdapter~\cite{li2023graphadapter} &   \underline{64.94} &  \underline{56.58}  & \underline{35.89}  & 23.07  & \underline{60.86}  & \underline{44.10}\\
\rowcolor{gray!20}
\textbf{SimNL (Ours)} &   \textbf{66.52}  & \textbf{57.87}  &  \textbf{36.38} & \underline{25.73}  &  \textbf{61.12} & \textbf{45.28}\\
\bottomrule
\end{tabular}
}
\vspace{-17pt}
\end{center}
\end{table}

\textbf{Robustness to Label Noise}. In Table~\ref{tab:label}, we report the performance of Tip-Adapter-F~\cite{zhang2022tip} and our SimNL on a noisy 16-shot ImageNet~\cite{deng2009imagenet}, where we randomly flip 10\% to 50\% labels. As shown, applying our instance reweighting
enhances both Tip-Adapter-F and SimNL's robustness to label noise.
Notably, our reweighting technique also improves performance when no label noise is introduced, as it can identify and downweight outliers during adaptation.

\begin{table}
\setlength\tabcolsep{10pt}
\caption{\looseness=-1 \textbf{Comparison of robustness to label noise on noisy 16-shot ImageNet~\cite{deng2009imagenet}}. We apply our instance reweighting technique to both Tip-Adapter-F~\cite{zhang2022tip} and our SimNL, and report the performance across four levels of noise.}
\vspace{-5pt}
\label{tab:label}
\resizebox{\linewidth}{!}{
\begin{tabular}{lcccc}
\toprule
Method & 0\% & 10\%  & 25\%   & 50\% \\ \midrule
Tip-Adapter-F~\cite{zhang2022tip} & 65.52 & 64.93 & 64.04 & 62.47 \\ 
 + Reweighting & \textbf{65.64} & \textbf{65.25} & \textbf{64.55} & \textbf{63.39} \\ \rowcolor{gray!20}
 \textit{Performance Gain} & \textcolor{MidnightBlue}{+0.12} & \textcolor{MidnightBlue}{+0.32} & \textcolor{MidnightBlue}{+0.51} &  \textcolor{MidnightBlue}{+0.92} \\ \midrule
SimNL (Ours) & 66.31 & 65.54 &  64.77 & 63.37  \\ 
 + Reweighting & \textbf{66.52} & \textbf{65.82} & \textbf{65.16} &  \textbf{64.02} \\ \rowcolor{gray!20}
\textit{Performance Gain} & \textcolor{MidnightBlue}{+0.21} & \textcolor{MidnightBlue}{+0.28} & \textcolor{MidnightBlue}{+0.39} &  \textcolor{MidnightBlue}{+0.65} \\
\bottomrule
\end{tabular}
}
\vspace{-12pt}
\end{table}

\begin{figure*}[t]
\centering
\includegraphics[width=\linewidth]{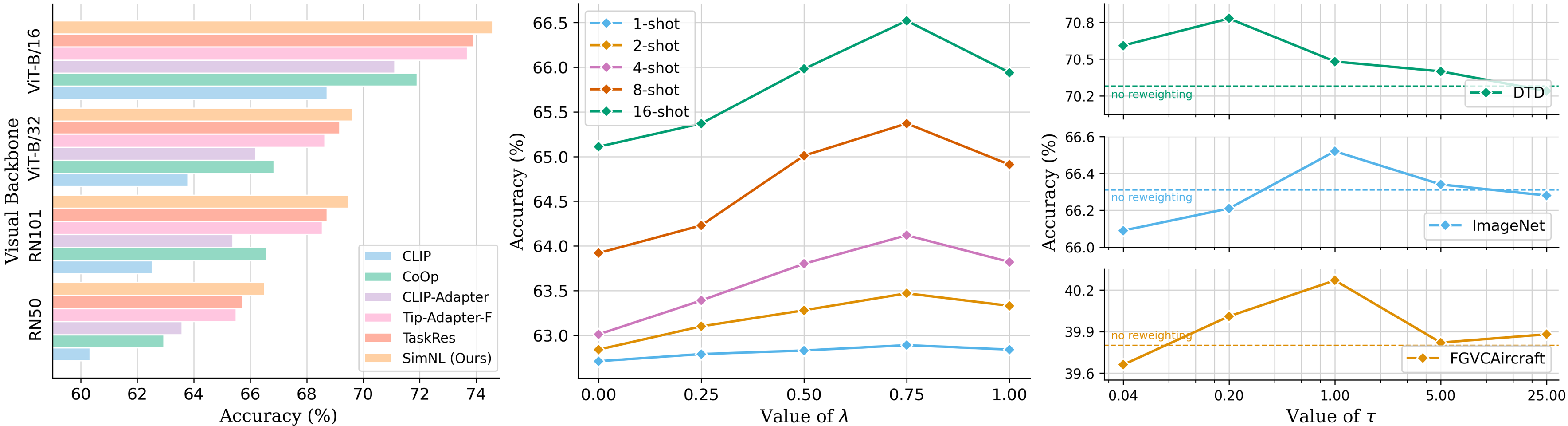}
\vspace{-18pt}
\caption{\looseness=-1 \textbf{More ablation results}. (\textit{Left}) Performance comparison of our SimNL with others on few-shot learning using different visual backbones; (\textit{Middle}) Sensitivity analysis of $\lambda$ from Eq.~(\ref{eq:lambda}) on ImageNet~\cite{deng2009imagenet}; (\textit{Right}) Sensitivity analysis of $\tau$  from Eq.~(\ref{eq:refine}) on 3 datasets.}
\label{fig:ablation}
\vspace{-8pt}
\end{figure*}

\textbf{Efficiency Comparison}. We also compare the efficiency of SimNL with existing methods in Table~\ref{tab:efficiency}. Our method achieves the highest accuracy while also exhibiting advantageous computational efficiency: (1) Compared to prompt-based learning methods such as CoOp~\cite{zhou2022learning} and ProGrad~\cite{zhu2023prompt}, our proposed method requires approximately 300$\times$ less training time and over 1000$\times$ fewer FLOPs since we do not need to propagate gradients through the textual encoder; (2) Compared to adapter-style fine-tuning methods, our SimNL requires 10$\times$ less training time than CLIP-Adapter~\cite{gao2021clip}, and demands 3$\times$ fewer FLOPs than Tip-Adapter-F~\cite{zhang2022tip} and GraphAdapter~\cite{li2023graphadapter}.

\begin{table}
\setlength\tabcolsep{6pt}
\caption{\textbf{Efficiency comparison with other existing methods on 16-shot ImageNet~\cite{deng2009imagenet}}. We report the training time, number of epochs, FLOPs, and the number of parameters for each method.}
\vspace{-5pt}
\label{tab:efficiency}
\resizebox{\linewidth}{!}{
\begin{tabular}{lccccc}
\toprule
Method & Training & Epochs  & GFLOPs   & Param. & Accuracy\\ \midrule
CLIP~\cite{radford2021learning} & - & - & - & - & 60.33 \\
CoOp~\cite{zhou2022learning} & 14 hr & 200 & $>$10 & \textbf{0.01M} & 62.95 \\
ProGrad~\cite{zhu2023prompt} & 17 hr & 200 & $>$10 & \textbf{0.01M} & 63.45 \\
CLIP-Adapter~\cite{gao2021clip} & 50 min & 200 & \textbf{0.004} & 0.52M & 63.59 \\
Tip-Adapter-F~\cite{zhang2022tip} & \textbf{5 min} & \textbf{20} & 0.030 & 16.38M & 65.51\\
GraphAdapter~\cite{li2023graphadapter} & \textbf{5 min} & \textbf{20} & 0.030 & 4.15M & 65.70\\
\cc \textbf{SimNL (Ours)} & \cc \textbf{5 min} & \cc \textbf{20} & \cc 0.009 & \cc 4.15M & \cc \textbf{66.52} \\
\bottomrule
\end{tabular}
}
\vspace{-5pt}
\end{table}

\subsection{Ablation Studies}

\noindent
\textbf{Effectiveness of Different Components}. 
In Table~\ref{table:ablation}, we conduct a systematic analysis of the impacts of various components within our SimNL framework. More specifically, we assess the performance of four distinct SimNL variants, each configured to allow two learnable residues to be updated while keeping the others fixed. 
We observe that the textual variant (SimNL-T) and the positive variant (SimNL-P) generally exhibit superior efficiency compared to their visual and negative counterparts. The full SimNL method, which integrates all four branches, surpasses the individual variants by achieving the highest average accuracy of 77.80\% in the 16-shot scenario.

\label{subsec:ablation}
\renewcommand\arraystretch{1.1}
\begin{table}[t]
\setlength\tabcolsep{4pt}
\caption{\looseness=-1 \textbf{Ablation studies for different variants of our method}. We evaluate the few-shot adaptation capabilities of four variants and report their average performance across all 11 datasets.}
\vspace{-5pt}
\label{table:ablation}
\centering
\resizebox{\linewidth}{!}{
\begin{tabular}{l|cccc|cccccc}
\toprule
 Method & $\mathcal{R}_{\mathsf{T}}^{+}$ & $\mathcal{R}_{\mathsf{T}}^{-}$ & $\mathcal{R}_{\mathsf{V}}^{+}$ & $\mathcal{R}_{\mathsf{V}}^{-}$ & 1-shot & 2-shot & 4-shot & 8-shot & 16-shot\\
\midrule
SimNL-T &\ding{51} & \ding{51} &  \ding{55}   & \ding{55} &   67.08   &  69.70  &  71.86  &  74.17 &  77.08  \\
SimNL-V &\ding{55} & \ding{55} &  \ding{51}   &  \ding{51} &   64.60  &  66.03   &  68.25  & 71.84 &  74.41  \\
SimNL-P &\ding{51} & \ding{55} &  \ding{51}   &  \ding{55} &  66.72  &  69.30   & 71.49   & 73.98  & 76.76  \\
SimNL-N &\ding{55} & \ding{51} &  \ding{55}   &  \ding{51} & 66.23   &  68.10   & 70.55  &  72.33   &  74.17  \\ 
\rowcolor{gray!20}
\textbf{SimNL} &\ding{51} & \ding{51} &  \ding{51}   &  \ding{51} &   \textbf{67.45}  & \textbf{70.06}  &  \textbf{72.43} & \textbf{75.06}  &  \textbf{77.80} \\
\bottomrule
\end{tabular}
}
\vspace{-10pt}
\end{table}

\textbf{Effects of Different Visual Backbones}. We also implement our SimNL with various visual encoders, including ResNet~\cite{he2016deep} and ViT~\cite{dosovitskiy2021an}, and evaluate their performance against other adaptation methods in Figure~\ref{fig:ablation} (\textit{Left}). We can see that our SimNL consistently exceeds other methods across all visual backbones, indicating the general effectiveness of our negative learning approach.

\looseness=-1
\textbf{Sensitivity Analysis of Hyper-Parameters}. We provide sensitivity analysis for the hyper-parameters $\lambda$ and $\tau$ in Figure~\ref{fig:ablation}. The hyper-parameter $\lambda$ from Eq.~(\ref{eq:lambda}) controls the combination of the positive and negative predictions. 
In Figure~\ref{fig:ablation} (\textit{Middle}), we can observe that setting $\lambda=0.75$ consistently yields the optimal performance. 
Moreover, in Figure~\ref{fig:ablation} (\textit{Right}), we adjust the temperature hyper-parameter $\tau$ in Eq.~(\ref{eq:refine}) from $0.04$ to $25$ and report the 16-shot accuracy of our method. We can see that by reweighting the few-shot samples, we achieve performance improvements ranging from 0.21\% to 0.55\% across three datasets. 

\looseness=-1
\textbf{Scalability to More Shots}. In Table~\ref{tab:scale}, we demonstrate that our SimNL scales effectively to more-shot settings, consistently outperforming Tip-Adapter-F~\cite{zhang2022tip} by 1.05\% to 1.24\% as the number of shots increases.

\label{subsec:scale}
\begin{table}
\setlength\tabcolsep{8pt}
\caption{\looseness=-1\textbf{Scalability to more shots}. We show performance comparison with Tip-Adapter(-F) on ImageNet at higher-shot settings.}
\vspace{-5pt}
\centering
\resizebox{\linewidth}{!}{
\begin{tabular}{lcccc}
\toprule
Method & 16-shot & 32-shot  & 64-shot   & 128-shot \\ \midrule
CLIP~\cite{radford2021learning} & \multicolumn{4}{c}{$^\dag$0-shot: 60.33}  \\
Tip-Adapter~\cite{zhang2022tip} & 62.03 & 62.51 & 62.88 & 63.15 \\
Tip-Adapter-F~\cite{zhang2022tip} & 65.47 & 66.58 & 67.96 & 69.74 \\
\cc \textbf{SimNL (Ours)} & \cc \textbf{66.52} & \cc \textbf{67.68} & \cc \textbf{69.01} & \cc \textbf{70.98} \\
\bottomrule
\end{tabular}
}
\vspace{-10pt}
\label{tab:scale}
\end{table}


\section{Conclusion}

\label{sec:conclusion}
\looseness=-1 
\noindent
In this work, we introduce SimNL, a simple and effective approach that applies the concept of negative learning to vision-language few-shot adaptation. Specifically, we transform the open-vocabulary CLIP model into a negative classifier, which is further optimized to exclude incorrect classes based on the image input.
During inference, we conduct simultaneous positive classification and negative exclusion to enhance the overall prediction accuracy.
Furthermore, we develop an unsupervised few-shot instance reweighting approach to mitigate the adverse effects of noisy image samples during few-shot adaptation .
Comprehensive evaluations on 15 diverse datasets demonstrate that our proposed SimNL outperforms the state-of-the-art methods in both few-shot learning and domain generalization tasks, while maintaining competitive efficiency.

\textbf{Acknowledgements}. This work has been funded in part by the Army Research Laboratory (ARL) award W911NF-23-2-0007, DARPA award FA8750-23-2-1015, and ONR award N00014-23-1-2840.

{\small
\bibliographystyle{ieee_fullname}
\bibliography{egbib}
}
\clearpage
\renewcommand{\thesection}{\Alph{section}}
\renewcommand\thefigure{\Alph{section}\arabic{figure}} 
\renewcommand\thetable{\Alph{section}\arabic{table}}  
\setcounter{section}{0}
\setcounter{figure}{0} 
\setcounter{table}{0} 

\maketitlesupplementary

In appendix, we provide additional details and experimental results to enhance understanding and insights into our proposed SimNL.
This supplementary document is organized as follows:
\begin{itemize}[leftmargin=0.5cm, itemindent=0cm, itemsep=4pt,topsep=4pt,parsep=0pt]
    \item[$\bullet$] \looseness=-1 Full numerical results for the few-shot learning task are detailed in Section~\ref{sec:full}.
    \item[$\bullet$] We compare our SimNL with other state-of-the-art methods in domain generalization tasks, utilizing an alternative ViT backbone, in Section~\ref{sec:moredg}.
    \item[$\bullet$] More sensitivity analyses of the hyper-parameters are conducted in Section~\ref{sec:hyper}.
    \item[$\bullet$] We differentiate our method from related work in Section~\ref{sec:related}.
    \item[$\bullet$] Detailed statistics for all utilized datasets are provided in Section~\ref{subsec:dataset}.
    \item[$\bullet$] We present the specific positive and negative prompts we used for each dataset in Section~\ref{sec:prompt}.
    \item[$\bullet$] We list the license information for all used assets in Section~\ref{sec:license}.
    \item[$\bullet$] Finally, we explore potential future work and discuss the limitations and broader impacts of this work in Section~\ref{sec:future}.
\end{itemize}

\vspace{-8pt}
\section{Additional Experimental Results}
\vspace{-2pt}
\subsection{Full Numerical Results on Few-Shot Learning}
\vspace{-5pt}
\label{sec:full}
In Figure~\ref{fig:fewshot} of the main text, we have evaluated our SimNL on few-shot learning task and compared with other state-of-the-art methods. In Table~\ref{tab:numerical_results}, we present the corresponding full numerical results on the few-shot learning task. We also report the 95\% confidence interval over 3 random seeds of our SimNL to ensure reliability of our results. In the last column, we present the average recognition accuracy over 11 datasets. The results indicate that our SimNL consistently outperforms other state-of-the-art methods across various few-shot learning settings by substantial margins.

Our SimNL demonstrates superior recognition performance across nearly all tested scenarios, with certain exceptions in lower-shot settings for the Food101~\cite{bossard2014food} and OxfordPets~\cite{parkhi2012cats} datasets. We attribute this to the prevalent challenge of overfitting, a common issue not exclusive to our approach but also affecting many existing methods, especially TaskRes~\cite{yu2023task}: While TaskRes consistently secures the second-best performance across the other 9 datasets, it underperforms significantly on these two. We hypothesize that this issue arises from the noisy training data with \textit{intense colors and sometimes wrong labels}~\cite{bossard2014food,parkhi2012cats}. However, in this work, we have designed a label refinement mechanism specifically to address this issue. As a result, our SimNL secures a relatively robust performance and achieves the second-best on these two datasets.

\begin{table*}[htpb]
\caption{\looseness=-1 \textbf{Full numerical results on few-shot learning task}. For each dataset, we report the mean accuracy and 95\% confidence interval over 3 random seeds of our SimNL on 1-/2-/4-/8-/16-shot settings. $^\dag$We report the zero-shot performance of CLIP~\cite{radford2021learning} for all settings. For TaskRes~\cite{yu2023task}, we report the results using the enhanced base classifier (\ie, TaskRes*). The best results are in \textbf{bold} and the second are \underline{underlined}.}
\vspace{-5pt}
\renewcommand\arraystretch{1.35}
\resizebox{\textwidth}{!}{
\begin{tabular}{l|c|ccccccccccccc|c}
\toprule
Method             & Setting                                           & \rotatebox{90}{Caltech101~\cite{fei2004learning}} & \rotatebox{90}{DTD~\cite{cimpoi2014describing}}   & \rotatebox{90}{EuroSAT~\cite{helber2019eurosat}} & \rotatebox{90}{FGVCAircraft~\cite{maji2013fine}} & \rotatebox{90}{Flowers102~\cite{nilsback2008automated}} & \rotatebox{90}{Food101~\cite{bossard2014food}} & \rotatebox{90}{ImageNet~\cite{deng2009imagenet}} & \rotatebox{90}{OxfordPets~\cite{parkhi2012cats}} & \rotatebox{90}{StanfordCars~\cite{krause20133d}} & \rotatebox{90}{SUN397~\cite{xiao2010sun}} & \rotatebox{90}{UCF101~\cite{soomro2012ucf101}} & Avg.  \\ \midrule
$^\dag$Zero-shot CLIP~\cite{radford2021learning}    & \multirow{8}{*}{1-shot}                                            & 84.52      & 40.33 & 41.80   & 16.98        & 65.46      & 77.31   & 60.33    & 85.51      & 54.26        & 58.56  & 61.44  & 58.77 \\
CoOp~\cite{zhou2022learning}                &                                                   & 87.43      & 44.13 & 50.51   & 9.80         & 67.90      & 73.71   & 57.15    & 86.51      & 55.48        & 60.10  & 62.10  & 59.53 \\
CoCoOp~\cite{zhou2022conditional}                &                                                   & 86.01      & 45.14 & 35.08   & 17.81         & 67.52      & 77.42   & 60.84   & \textbf{86.96}      & 57.22        & 62.28  & 62.84  & 59.92 \\
CLIP-Adapter~\cite{gao2021clip}        &                                                   & 88.70      & 45.66 & 61.51   & 17.21        & 73.43      & 76.77   & 61.20    & 85.99      & 55.14        & 61.28  & 62.29  & 62.65 \\
Tip-Adapter-F~\cite{zhang2022tip}       &                                                   & \underline{89.38}      & \underline{50.31} & 59.16   & 20.83        & \underline{80.13}      & \underline{77.61}   & 61.32    & 86.47    & 58.51        & \underline{62.51}  & \underline{64.91}  & \underline{64.65} \\
TaskRes~\cite{yu2023task}             &                                                   & 88.80      & 50.20 & \underline{61.70}   & \underline{21.41}        & 79.17      & 74.03   & \underline{61.90}    & 83.60      & \underline{59.13}        & 62.33  & 64.77  & 64.28 \\ \rowcolor{gray!20}
&                                                   & \textbf{90.87}      & \textbf{53.13} & \textbf{67.70}   & \textbf{23.61}        & \textbf{84.17}      & \textbf{77.93}   & \textbf{62.89}    & \underline{86.90}      & \textbf{61.34}        & \textbf{65.13}  & \textbf{68.25}  & \textbf{67.45} \\ \rowcolor{gray!20}
\multirow{-2}{*}[-0.1em]{SimNL (Ours)}   & &  {\small ($\pm 0.33$)}  & {\small ($\pm 0.48$)}  & {\small ($\pm 0.47$)} &  {\small ($\pm 0.18$)} & {\small ($\pm 0.34$)} & {\small ($\pm 0.22$)} & {\small ($\pm 0.13$)} & {\small ($\pm 0.31$)} &  {\small ($\pm 0.28$)} & {\small ($\pm 0.08$)} & {\small ($\pm 0.23$)} & {\small ($\pm 0.28$)}\\ \midrule
$^\dag$Zero-shot CLIP~\cite{radford2021learning}      &   \multirow{8}{*}{2-shot}                        & 84.52      & 40.33 & 41.80   & 16.98        & 65.46      & 77.31   & 60.33    & 85.51      & 54.26        & 58.56  & 61.44  & 58.77 \\
CoOp~\cite{zhou2022learning}                &                           & 87.92      & 45.04 & 60.43   & 18.25        & 77.47      & 72.26   & 55.88    & 82.36      & 58.10        & 59.82  & 64.13  & 61.97 \\
CoCoOp~\cite{zhou2022conditional}                &                                                   & 89.47     & 46.20 & 38.51   & 20.22         & 70.70      & \textbf{78.81}   & 61.86    & \textbf{88.81}      & 58.28        & 63.50  & 65.23  & 61.96 \\
CLIP-Adapter~\cite{gao2021clip}        &                          & 89.32      & 51.81 & 64.11   & 20.10        & 81.77      & 77.20   & 61.52    & 86.73      & 58.71        & 62.21  & 67.27  & 65.52 \\
Tip-Adapter-F~\cite{zhang2022tip}       &                           & 89.81      & 54.00 & 65.82   & 23.47        & 82.50      & 77.83   & 61.69    & 87.10      & 62.05        & 63.55  & 66.23  & 66.73 \\
TaskRes~\cite{yu2023task}             &                          & \underline{90.27}      & \underline{55.13} & \underline{65.83}   & \underline{24.13}        & \underline{86.57}      & 75.17   & \underline{62.63}    & 84.63      & \underline{63.70}        & \underline{64.97}  & \underline{70.00}  & \underline{67.54} \\ \rowcolor{gray!20}
&   & \textbf{91.19}    & \textbf{59.17} & \textbf{74.40}   & \textbf{26.22}        & \textbf{88.43}      & \underline{78.35}   & \textbf{63.47}    & \underline{87.68}      & \textbf{64.77}        & \textbf{66.73}  & \textbf{70.25}  & \textbf{70.06} \\ \rowcolor{gray!20}
\multirow{-2}{*}[-0.1em]{SimNL (Ours)}  &  & {\small ($\pm 0.26$)} & {\small ($\pm 0.31$)} & {\small ($\pm 0.87$)} &  {\small ($\pm 0.37$)} & {\small ($\pm 0.18$)} & {\small ($\pm 0.17$)} & {\small ($\pm 0.06$)} & {\small ($\pm 0.28$)} &  {\small ($\pm 0.21$)} & {\small ($\pm 0.33$)} & {\small ($\pm 0.48$)} & {\small ($\pm 0.31$)}\\

\midrule
$^\dag$Zero-shot CLIP~\cite{radford2021learning}      &       \multirow{8}{*}{4-shot}                   & 84.52      & 40.33 & 41.80   & 16.98        & 65.46      & 77.31   & 60.33    & 85.51      & 54.26        & 58.56  & 61.44  & 58.77 \\
CoOp~\cite{zhou2022learning}                &                         & 89.17      & 53.38 & 70.20   & 21.72        & 85.81      & 72.72   & 59.93    & 87.22      & 61.92        & 63.46  & 67.08  & 66.60 \\
CoCoOp~\cite{zhou2022conditional}                &                                                   & 90.31      & 47.90 & 63.56   & 20.56         & 72.72      & \textbf{79.51}   & 62.52    & \textbf{88.60}      & 59.90        & 64.90  & 67.90  & 65.31 \\
CLIP-Adapter~\cite{gao2021clip}        &                           & 89.98      & 57.02 & 73.18   & 22.99        & 87.30      & 77.93   & 61.84    & 87.36      & 62.26        & 65.90  & 68.90  & 68.61 \\
Tip-Adapter-F~\cite{zhang2022tip}       &                        & 90.67      & 57.78 & \underline{73.85}   & \underline{26.01}        & 89.02      & 78.26   & 62.52    & 87.72      & 64.82        & 66.13  & 70.87  & 69.79 \\
TaskRes~\cite{yu2023task}             &                          & \underline{90.97}      & \underline{60.70} & 73.83   & 25.70        & \underline{90.20}      & 76.10   & \underline{63.57}    & 86.33      & \underline{67.43}        & \underline{67.27}  & \underline{70.93}  & \underline{70.28} \\ \rowcolor{gray!20}
 &  &  \textbf{92.21}      & \textbf{66.01} & \textbf{76.54}   & \textbf{28.95}        & \textbf{92.04}      & \underline{78.74}   & \textbf{64.12}    & \underline{88.13}      & \textbf{67.96}        & \textbf{68.59}  & \textbf{73.46}  & \textbf{72.43} \\ \rowcolor{gray!20}
\multirow{-2}{*}[-0.1em]{SimNL (Ours)}  &  &  {\small ($\pm 0.21$)} & {\small ($\pm 0.49$)} & {\small ($\pm 0.59$)}  &  {\small ($\pm 0.29$)} & {\small ($\pm 0.25$)} & {\small ($\pm 0.11$)} & {\small ($\pm 0.17$)} & {\small ($\pm 0.26$)} &  {\small ($\pm 0.36$)} & {\small ($\pm 0.29$)} &  {\small ($\pm 0.51$)} & {\small ($\pm 0.32$)} \\
\midrule
$^\dag$Zero-shot CLIP~\cite{radford2021learning}      & \multirow{8}{*}{8-shot}           & 84.52      & 40.33 & 41.80   & 16.98        & 65.46      & 77.31   & 60.33    & 85.51      & 54.26        & 58.56  & 61.44  & 58.77 \\
CoOp~\cite{zhou2022learning}                &                          & 90.15      & 59.88 & 76.51   & 25.93        & 90.84      & 71.52   & 60.91    & 86.40      & 68.49        & 65.63  & 71.81  & 69.82 \\
CoCoOp~\cite{zhou2022conditional}                &                                                   & 90.14      & 52.21 & 64.13   & 22.03         & 75.88      & \textbf{79.59}   & 62.40    & \underline{88.74}      & 60.87        & 65.37  & 68.25  & 66.33 \\
CLIP-Adapter~\cite{gao2021clip}        &                          & 91.22      & 60.70 & 78.34   & 25.77        & 91.79     & 78.01   & 62.68    & 87.70      & 67.78        & 67.52  & 73.02  & 71.32 \\
Tip-Adapter-F~\cite{zhang2022tip}       &                         & 91.54      & 62.67 & 77.83   & 30.21        & 91.85      & 78.71   & 64.00    & 88.07      & 69.53        & \underline{68.80}  & 74.50  & 72.52 \\
TaskRes~\cite{yu2023task}             &                         & \underline{92.40}      & \underline{64.77} & \underline{79.33}   & \underline{31.48}        & \underline{94.73}      & 76.40   & \underline{64.67}    & 87.17      & \underline{71.83}        & 68.73  & \underline{75.33}  & \underline{73.35} \\ \rowcolor{gray!20}
 &   & \textbf{93.40}     & \textbf{67.78} & \textbf{81.62}   & \textbf{33.90}        & \textbf{95.23}      & \underline{79.23}   & \textbf{65.37}    & \textbf{89.29}     & \textbf{72.08}        & \textbf{70.93}  & \textbf{76.84}  & \textbf{75.06} \\ \rowcolor{gray!20}
\multirow{-2}{*}[-0.1em]{SimNL (Ours)} & & {\small ($\pm 0.23$)} & {\small ($\pm 0.55$)} & {\small ($\pm 0.54$)} &  {\small ($\pm 0.63$)} & {\small ($\pm 0.26$)} & {\small ($\pm 0.21$)} & {\small ($\pm 0.09$)} & {\small ($\pm 0.21$)} &  {\small ($\pm 0.51$)} & {\small ($\pm 0.15$)} & {\small ($\pm 0.39$)} & {\small ($\pm 0.34$)}  \\
\midrule
$^\dag$Zero-shot CLIP~\cite{radford2021learning}      &      \multirow{8}{*}{16-shot}                    & 84.52      & 40.33 & 41.80   & 16.98        & 65.46      & 77.31   & 60.33    & 85.51      & 54.26        & 58.56  & 61.44  & 58.77 \\
CoOp~\cite{zhou2022learning}                &                          & 91.61      & 63.11 & 82.36   & 31.01        & 94.39      & 73.80   & 62.95    & 87.30      & 72.51        & 69.11  & 75.70  & 73.07 \\
CoCoOp~\cite{zhou2022conditional}                &                                                   & 90.90      & 57.53 & 70.77   & 22.40         & 79.14      & \underline{79.68}   & 62.71    & \underline{89.93}      & 62.22       & 67.21  & 70.81  & 68.48 \\
CLIP-Adapter~\cite{gao2021clip}        &                        & 92.44      & 66.14 & 82.76   & 31.83        & 93.91      & 78.21   & 63.59    & 87.91      & 74.12        & 69.59  & 76.80  & 74.30 \\
Tip-Adapter-F~\cite{zhang2022tip}       &                          & 92.93      & \underline{67.33} & 83.80   & 35.50        & 95.01      & 79.50   & 65.51    & 89.71      & 75.50        & \underline{71.31}  & \underline{78.01}  & \underline{75.83} \\
TaskRes~\cite{yu2023task}             &                          & \underline{93.43}      & 67.13 & \underline{84.03}   & \underline{36.30}        & \underline{96.03}      & 77.60   & \underline{65.73}    & 87.83      & \underline{76.83}        & 70.67  & 77.97  & 75.78 \\ \rowcolor{gray!20}
  & & \textbf{93.77}       & \textbf{70.83} & \textbf{87.36}   & \textbf{40.27}        & \textbf{96.51}      & \textbf{79.87}   & \textbf{66.52}    & \textbf{90.58}      & \textbf{77.48}       & \textbf{72.32}  & \textbf{80.28}  & \textbf{77.80} \\  \rowcolor{gray!20}
\multirow{-2}{*}[-0.1em]{SimNL (Ours)} &  &  {\small ($\pm 0.33$)} & {\small ($\pm 0.77$)}  &  {\small ($\pm 0.84$)} &  {\small ($\pm 0.53$)} & {\small ($\pm 0.58$)} & {\small ($\pm 0.34$)} & {\small ($\pm 0.13$)} & {\small ($\pm 0.38$)} &  {\small ($\pm 0.68$)} & {\small ($\pm 0.24$)} & {\small ($\pm 0.26$)} & {\small ($\pm 0.50$)}  \\
\bottomrule
\end{tabular}}
\label{tab:numerical_results}
\end{table*}

\subsection{More Results on Domain Generalization}
\begin{table}
\small
\begin{center}
\caption{\looseness=-1 \textbf{Performance comparison on robustness to distribution shifts}. All the models are trained on 16-shot ImageNet~\cite{deng2009imagenet} and directed tested on the OOD target datasets. The best results are in \textbf{bold} and the second best are \underline{underlined}.}
\label{tab:moredg}
\resizebox{\linewidth}{!}{
\begin{tabular}{lcccccc}
\toprule
\multirow{2}{*}[-0.5ex]{Method}  & Source & \multicolumn{5}{c}{Target} \\ \cmidrule(lr){2-2} \cmidrule(lr){3-7}  & ImageNet & -V2 & -Sketch & -A & -R  & Avg. \\
\midrule
Zero-Shot CLIP~\cite{radford2021learning}  &  62.05 &  54.79  & 40.82  & 29.57 &  65.99  & 47.79\\
Linear Probe CLIP~\cite{radford2021learning}  & 59.58  & 49.73  & 28.06  & 19.67  & 47.20 &  36.17\\
CoOp~\cite{zhou2022learning}   & 66.85  & 58.08  & 40.44  & 30.62  & 64.45 &  48.40\\
TaskRes~\cite{yu2023task} &   68.20 &  \underline{59.20}  & 42.50  & 31.43  & 69.33  & 50.62\\
GraphAdapter~\cite{li2023graphadapter} &   \underline{68.47} &  59.10  & \underline{42.70}  & \underline{31.73}  & \underline{69.43}  & \underline{50.74}\\
\rowcolor{gray!20}
SimNL (Ours) &   \textbf{69.63}  & \textbf{59.76}  &  \textbf{43.41} & \textbf{32.48}  &  \textbf{69.60} & \textbf{51.31}\\
\bottomrule
\end{tabular}
}
\end{center}
\end{table}
In Table~\ref{table:generalization} in Section~\ref{subsec:results} of the main text, we compare the generalization performance of our SimNL with other state-of-the-art methods in the presence of distribution shifts with ResNet-50 visual backbone. In Table~\ref{tab:moredg}, we present the performance comparison on domain generalization task using ViT-B/32 visual backbone. Consistently, our SimNL not only achieves state-of-the-art performance on the source dataset but also attains an average performance gain of 0.57\% across 4 out-of-distribution (OOD) target datasets. This verifies that our SimNL demonstrates superior generalizability compared to other state-of-the-art methods, independent of the visual backbone utilized.

\label{sec:moredg}

\subsection{More Sensitivity Analyses of Hyper-Parameters}
\label{sec:hyper}
\begin{table}
\caption{\textbf{Sensitivity of hyper-parameters}. All the results are reported on 16-shot ImageNet~\cite{deng2009imagenet}.}
\centering
\vspace{2pt}
\begin{adjustbox}{width=\linewidth}
\setlength{\tabcolsep}{3mm}{
	\begin{tabular}{c|cccccc}
	\toprule
		\multirow{2}{*}[-0.1em]{$\alpha$}  &0.0 &0.5 &1.0 &\textbf{1.2} &1.5 &2.0 \\  
        \cmidrule(lr){2-7}
		 & 66.14 & 66.30  & 66.41 & \textbf{66.52} & 66.44 & 66.36\\ 
        \midrule
	    \multirow{2}{*}[-0.1em]{$\beta$} &1.0 &1.5 &\textbf{2.0} &2.5 &3.0 &3.5 \\
	     \cmidrule(lr){2-7}
	    & 66.38  & 66.44& \textbf{66.52} & 66.50 &66.48 &66.40\\   
	\bottomrule
	\end{tabular}
} 
\end{adjustbox}
\label{tab:hyper}
\end{table}
\looseness=-1
Building upon the sensitivity analyses of $\lambda$ and $\tau$ detailed in Section~\ref{subsec:ablation} of the main text, this section extends our examination to include the sensitivity of parameters $\alpha$ and $\beta$ on 16-shot ImageNet~\cite{deng2009imagenet}.
In our experiments on ImageNet~\cite{deng2009imagenet}, we set the hyper-parameters $\alpha$ and $\beta$ defined in Section~\ref{sec:method} to $1.2$ and $2.0$, respectively. To comprehensively investigate the effects of different hyper-parameters, we conducted a sensitivity experiment where we varied each hyper-parameter individually and evaluated the performance on 16-shot ImageNet~\cite{deng2009imagenet} in Fig.~\ref{tab:hyper}. We can see that our choice of $\alpha=1.2$ and $\beta=2.0$ yields the highest performance. Moreover, our SimNL maintains robust performance when adjusting these two hyper-parameters, since our SimNL includes adapters from both textual and visual modalities and each of our four adapters can work effectively, as we presented in Table~\ref{table:ablation} in the main text.

\section{More Discussions on Related Work}
\label{sec:related}
\subsection{Differences between Our Proposed Method and Traditional Negative Learning}
\label{subsec:difference2}
In this work, we apply the concept of negative learning to vision-language few-shot adaptation. However, our method is different with traditional negative learning approaches~\cite{ishida2017learning,yu2018learning}. Specifically, traditional negative learning approaches aim to learn a positive classifier using negatibe labels, while our method aim to learn a negative classifier using the positive label.

\textbf{Negative Learning}. In the negative learning setting~\cite{ishida2017learning,yu2018learning}, we are given complementary labels $\bar{y} \in \mathcal{Y}\, \backslash \{y\}$, which represents the image does not belong to a specific class. Similarly, the encoded one-hot labels are given by $\bar{\boldsymbol{y}} \in \{0, 1\}^C$. This leads to a complementary cross-entropy loss for optimizing the classifier parameters $\theta$. The training objective for $\mathcal{F}_{\theta}$ can thus be expressed as
\begin{gather}
    \min_{\theta} \mathcal{R}(\mathcal{F}_{\theta})= \mathbb{E}_{(\boldsymbol{x}, \bar{y}) \sim P(\boldsymbol{x}, \bar{y})} \left[\mathcal{L}\left(\mathcal{F}_{\theta}(\boldsymbol{x}), \bar{y} \right)\right], \nonumber\\ \text{where}\,\, \mathcal{L}\left(\mathcal{F}_{\theta}(\boldsymbol{x}), \bar{y} \right) = - \sum_{k=1}^{C} \bar{\boldsymbol{y}}_k \log  (1 -\boldsymbol{p}_k).
\end{gather}
Here, the goal is to minimize the probability corresponding to the complementary label, \ie, $\boldsymbol{p}_{\bar{y}} \rightarrow 0$.

\textbf{Our Method}. In this work, we apply the concept of negative learning to vision-language few-shot adaptation by employing a distinct CLIP-based negative classifier $\mathcal{G}_{\varphi}: \mathcal{X} \rightarrow \mathbb{R}^C$ with parameters $\varphi$. This classifier predicts the negative probability $\bar{\boldsymbol{p}} = \texttt{Softmax}(\mathcal{G}_{\varphi}(\boldsymbol{x}))$ that the image does not belong to specific classes. The expected classification risk for  $\mathcal{G}_{\varphi}$ can be written as
\begin{gather}
    \min_{\varphi} \mathcal{R}(\mathcal{G}_{\varphi})= \mathbb{E}_{(\boldsymbol{x}, y) \sim P(\boldsymbol{x}, y)} \left[\mathcal{L}\left(\mathcal{G}_{\varphi}(\boldsymbol{x}), y \right)\right], \nonumber\\ \text{where}\,\, \mathcal{L}\left(\mathcal{G}_{\varphi}(\boldsymbol{x}), y \right) = - \sum_{k=1}^{C} \boldsymbol{y}_k \log (1 - \bar{\boldsymbol{p}}_k).
\end{gather}
By optimizing this risk, we aim to reduce the negative probability $\bar{\boldsymbol{p}}_{y} \rightarrow 0$ for the true label. 

\subsection{Differences between Our Proposed Method and Contrastive Learning}
\label{subsec:difference}
In this work, we propose a negative learning-based approach for vision-language few-shot adaptation. However, this ``negative" refers to a negative classifier, not the negative samples pairs in contrastive learning literature. In the following discussions, we show that our negative learning is fundamentally different with contrastive learning.

Specifically, CLIP operates as a similarity-based classifier where classification logits are derived from the similarities between image features $f_{v}$ and class-specific text features $\{f_{t_c}\}_{c=1}^C$:
\begin{equation}
\boldsymbol{p}_y = \frac{\exp \left( \mathrm{cos}\left(f_{t_y},f_{v} \right) /t \right)}{\sum\nolimits_{t'}{\exp \left( \mathrm{cos}\left( f_{t'},f_{v} \right) /t \right)}}, 
\end{equation}
In the context of few-shot adaptation of CLIP, both prompt-based learning methods and adapter-style fine-tuning methods seek to optimize $\{f_{t_c}\}_{c=1}^C$, whether by tuning the input prompt or directly modulating these features. As discussed in Section~\ref{sec:formulation}, these methods primarily employ positive learning, updating parameters via cross-entropy loss:
\begin{align}
\label{eq:contrastive}
    \mathcal{L}\left(\mathcal{F}_{\theta}(\boldsymbol{x}), y \right) &= - \sum_{k=1}^{C} \boldsymbol{y}_k \log \boldsymbol{p}_k \nonumber\\&= - \log \frac{\exp \left( \mathrm{cos}\left(f_{t_y},f_{v} \right) /t \right)}{\sum\nolimits_{t'}{\exp \left( \mathrm{cos}\left( f_{t'},f_{v} \right) /t \right)}},
\end{align}
where $y$ is the ground-truth label for sample $\boldsymbol{x}$. This update strategy aims to optimize $p_y=1$. 
By considering $f_{t_y}$ and $f_v$ as a positive pair and $f_{t_{\neq y}}$ and $f_v$ as negative pairs, the positive learning process essentially becomes contrastive in nature. Specifically,  Eq.~(\ref{eq:contrastive}) has the same form with the contrastive InfoNCE loss~\cite{oord2018representation}, which seeks to minimize the distance between the positive pair $f_{t_y}$ and $f_v$ while maximizing the distance between the negative pairs $f_{t_{\neq y}}$ and $f_v$.

However, in this work, we apply negative learning and introduce another negative classifier $\mathcal{G}_{\varphi}$. Specifically, we mine a general negative feature $f_{t_c}^-$ for each class $c$, which is absent in samples from class $c$ but present in samples from all other classes. Then the probability that the image not belongs to class $y$ can be written as
\begin{equation}
    \bar{\boldsymbol{p}}_y = \frac{\exp \left( \mathrm{cos}\left(f_{t_y}^-,f_{v} \right) /t \right)}{\sum\nolimits_{t'}{\exp \left( \mathrm{cos}\left( f_{t'}^-,f_{v} \right) /t \right)}}
\end{equation}
As discussed in Section~\ref{sec:formulation}, we update the set of negative features by
\begin{align}
\label{eq:contrastive2}
    \mathcal{L}\left(\mathcal{G}_{\varphi}(\boldsymbol{x}), y \right) &= - \sum_{k=1}^{C} \boldsymbol{y}_k \log (1 - \bar{\boldsymbol{p}}_k) \nonumber\\ &= - \log \left(1-  \frac{\exp \left( \mathrm{cos}\left(f_{t_y}^-,f_{v} \right) /t \right)}{\sum\nolimits_{t'}{\exp \left( \mathrm{cos}\left( f_{t'}^-,f_{v} \right) /t \right)}}\right).
\end{align}
In the language of contrastive learning, we consider all $f_{t_{\neq y}}^-$ and $f_{v}$ as positive pairs and $f_{t_{y}}^-$ and $f_{v}$ as the only negative pair.

In summary, the concept of ``negative" is different in our negative learning and contrastive learning: (1) In our negative learning, ``negative" specifically refers to a negative classifier. We explicitly train another negative classifier $\mathcal{G}_{\varphi}$ to ensemble with the positive classifier; (2) In contrastive learning, ``negative" refers to the negative sample pairs that are constructed and utilized during training. Specifically, by constructing positive and negative sample pairs, the objective of each classifier can be formulated as a contrastive objective as discussed above.

\subsection{Differences with Other Similar Approaches}
To the best of our knowledge, the following works adopt similar negative learning methods to open-set problems:
\begin{itemize}
    \item \textbf{RPL}~\cite{chen2020learning} is the seminal work that learns a set of reciprocal points as negative representations of target classes to enhance model's recognition capabilities when handling unseen samples.
    \item \textbf{CLIPN}~\cite{wang2023clipn} leverages additional large-scale datasets to a fine-tune a ``no" text encoder to enhance CLIP's out-of-distribution (OOD) detection capability.
    \item \textbf{LSN}~\cite{nie2024outofdistribution} also focus on learning some negative prompts for OOD detection task using CLIP.
\end{itemize}
However, we formulate negative learning from a different and more fundamental perspective, \ie, using negative prediction to directly improve the accuracy of positive prediction in a closed-set problem. We also empirically validate that leveraging negative cues from CLIP can effectively improve both few-shot classification performance and generalization capability.


\section{Additional Technical Details}

\subsection{Dataset Details}
\label{subsec:dataset}
In Table~\ref{tab:dataset}, we present the detailed statistics of each dataset we used in our experiments, including the number of classes, the sizes of training, validation and testing sets, and their original tasks.

\begin{table}[t]
\renewcommand\arraystretch{1.1}
    \caption{\looseness=-1 \textbf{Detailed statistics of datasets used in experiments}. Note that the last 4 ImageNet variant datasets are designed for evaluation and only contain the test sets.}
    \vspace{-5pt}
    \label{tab:dataset}
    \resizebox{\linewidth}{!}{
    \setlength{\tabcolsep}{1.5mm}{
    \begin{tabular}{lccccc}
    \toprule
Dataset                  & Classes  & Training & Validation   & Testing & Task \\ \midrule
Caltech101~\cite{fei2004learning} & 100 & 4,128 & 1,649 & 2,465& Object recognition \\
DTD~\cite{cimpoi2014describing}& 47 & 2,820 & 1,128& 1,692 &  Texture recognition\\ 
EuroSAT~\cite{helber2019eurosat}& 10 & 13,500 & 5,400& 8,100 & Satellite image recognition \\ 
FGVCAircraft~\cite{maji2013fine} & 100 & 3,334 & 3,333& 3,333 & Fine-grained aircraft recognition\\
Flowers102~\cite{nilsback2008automated} & 102 & 4,093 & 1,633& 2,463 & Fine-grained flowers recognition \\ 
Food101~\cite{bossard2014food} & 101 & 50,500& 20,200& 30,300 & Fine-grained food recognition  \\ 
ImageNet~\cite{deng2009imagenet} & 1,000 & 1.28M & -& 50,000 & Object recognition \\ 
OxfordPets~\cite{parkhi2012cats} & 37  & 2,944 & 736& 3,669 & Fine-grained pets recognition \\ 
StanfordCars~\cite{krause20133d} & 196 & 6,509 & 1,635& 8,041 & Fine-grained car recognition \\
SUN397~\cite{xiao2010sun}& 397& 15,880 & 3,970& 19,850 & Scene recognition\\ 
UCF101~\cite{soomro2012ucf101}& 101 & 7,639 & 1,898& 3,783 & Action recognition\\
\midrule
ImageNet-V2~\cite{recht2019imagenet} & 1,000 & - & -& 10,000 & Robustness of collocation  \\
ImageNet-Sketch~\cite{wang2019learning} & 1,000 & - & -&50,889 & Robustness of sketch domain\\
ImageNet-A~\cite{hendrycks2021natural}& 200 & - & -&7,500 &Robustness of adversarial attack\\
ImageNet-R~\cite{hendrycks2021many}& 200 & - & -&30,000&Robustness of multi-domains\\
    \bottomrule
    \end{tabular}
    }
    }
\end{table}

\subsection{Positive and Negative Prompts}
\label{sec:prompt}
In Table~\ref{tab:prompt}, we detail the specific positive and negative prompts utilized for each dataset. Additionally, as mentioned in Section 4.1, we incorporate prompts from CuPL~\cite{pratt2023does} to further enhance model performance.

\begin{table}[t]
\renewcommand\arraystretch{1.2}
    \caption{\looseness=-1 \textbf{Positive and negative prompts used in experiments}. In addition to these prompts, we also employ CuPL~\cite{pratt2023does} prompts to further enhance performance.}
    \vspace{-5pt}
    \label{tab:prompt}
    \resizebox{\linewidth}{!}{
    \setlength{\tabcolsep}{3mm}{
    \begin{tabular}{lccccc}
    \toprule
Dataset                  & Positive Prompts  & Negative Prompts  \\ \midrule
& ``itap of a \{\texttt{CLASS}\}.'' & ``itap without any \{\texttt{CLASS}\}.''\\ 
ImageNet~\cite{deng2009imagenet}& ``a bad photo of the \{\texttt{CLASS}\}.'' & ``a bad photo with no \{\texttt{CLASS}\} in it.''\\ 
ImageNet-V2~\cite{recht2019imagenet}& ``a origami \{\texttt{CLASS}\}.'' & ``a origami that isn't a \{\texttt{CLASS}\}.''\\ 
ImageNet-Sketch~\cite{wang2019learning}& ``a photo of the large \{\texttt{CLASS}\}.'' & ``a photo with no large \{\texttt{CLASS}\}.''\\ 
ImageNet-A~\cite{hendrycks2021natural}& ``a \{\texttt{CLASS}\} in a video game.'' & ``a video game scene without a \{\texttt{CLASS}\}.''\\ 
ImageNet-R~\cite{hendrycks2021many}& ``art of the \{\texttt{CLASS}\}.'' & ``art that doesn't include a \{\texttt{CLASS}\}.''\\ 
& ``a photo of the small \{\texttt{CLASS}\}.'' & ``a photo with no small \{\texttt{CLASS}\}.''\\ \midrule
Caltech101~\cite{fei2004learning} & ``a photo of a \{\texttt{CLASS}\}.'' & ``a photo without \{\texttt{CLASS}\}.'' \\
DTD~\cite{cimpoi2014describing}& ``\{\texttt{CLASS}\} texture.'' & ``not \{\texttt{CLASS}\} texture.''\\ 
EuroSAT~\cite{helber2019eurosat}& ``a centered satellite photo of \{\texttt{CLASS}\}.'' & ``a centered satellite photo without \{\texttt{CLASS}\}.'' \\ 
FGVCAircraft~\cite{maji2013fine} & ``a photo of a \{\texttt{CLASS}\}, a type of aircraft.'' & ``a photo without \{\texttt{CLASS}\}, a type of aircraft.''\\
Flowers102~\cite{nilsback2008automated} & ``a photo of a \{\texttt{CLASS}\}, a type of flower.'' & ``a photo without \{\texttt{CLASS}\}, a type of flower.'' \\ 
Food101~\cite{bossard2014food} & ``a photo of \{\texttt{CLASS}\}, a type of food.'' & ``a photo without \{\texttt{CLASS}\}, a type of food.''\\ 
OxfordPets~\cite{parkhi2012cats} & ``a photo of a \{\texttt{CLASS}\}, a type of pet.''  & ``a photo without \{\texttt{CLASS}\}, a type of pet.''\\ 
StanfordCars~\cite{krause20133d} & ``a photo of a \{\texttt{CLASS}\}.'' & ``a photo of no \{\texttt{CLASS}\}.'' \\
SUN397~\cite{xiao2010sun}& ``a photo of a \{\texttt{CLASS}\}.''& ``a photo without \{\texttt{CLASS}\}.'' \\ 
UCF101~\cite{soomro2012ucf101}& ``a photo of a person doing \{\texttt{CLASS}\}.'' & ``a photo of a person not doing \{\texttt{CLASS}\}.'' \\
    \bottomrule
    \end{tabular}
    }
    }
\end{table}

\section{License Information}
\label{sec:license}
\textbf{Datasets}. We list the known license information for the datasets below:
\begin{itemize}
    \item MIT License: ImageNet-A~\cite{hendrycks2021natural}, ImageNet-V2~\cite{recht2019imagenet}, ImageNet-R~\cite{hendrycks2021many}, and ImageNet-Sketch~\cite{wang2019learning}.
    \item CC BY-SA 4.0 License: OxfordPets~\cite{parkhi2012cats}.
    \item Research purposes only: ImageNet~\cite{deng2009imagenet}, StandfordCars~\cite{krause20133d}, DTD~\cite{cimpoi2014describing}, FGVCAircraft~\cite{maji2013fine}, SUN397~\cite{xiao2010sun}.
\end{itemize}

\textbf{Code}. In this work, we also use some code implementations from existing codebase: CLIP~\cite{radford2021learning}, CoOp~\cite{zhou2022learning}, APE~\cite{zhu2023not}, and CuPL~\cite{pratt2023does}. The code used in this paper are all under the MIT License.

\section{Further Discussions}
\label{sec:future}

\textbf{Future Work}. In this work, we introduce negative learning for adapting VLMs to downstream tasks.
We believe that the similar concept of \textit{negative} learning can also be applied to prompt-based learning methods, and be extended to fine-tune other foundational models (\eg, other VLMs~\cite{jia2021scaling,yu2022coca} and LLMs~\cite{touvron2023llama,devlin2019bert}). Besides, we also notice that there are a lot of research efforts dedicated to design better prompts (\eg, using LLM~\cite{pratt2023does,maniparambil2023enhancing,wu2023gpt4vis}) to fully exploit the capabilities of CLIP. We hope that with our work, future research endeavors can also be directed to investigate the utilization of negative prompts to better activate the negative inference capabilities, further broadening the scope and effectiveness of CLIP.

\textbf{Limitations}. We identify two potential limitations of our SimNL: (1) Its efficacy in zero-shot situations is constrained due to the scarcity of original negative descriptions in the CLIP training corpus; (2) Similar to other adapter-style fine-tuning approaches, our SimNL fine-tuned on a specific task cannot be directly applied to another task without additional adaptation. However, recently, Wang \etal~\cite{wang2024a} shows that adapter-style fine-tuning methods can be extended for these scenarios using the $k$NN algorithm.

\textbf{Broader Impacts.} In this work, we aim to build more reliable machine learning systems by leveraging the extensive knowledge of current foundational models. Specifically, we introduce negative learning to more efficiently transfer pre-trained VLMs to downstream tasks, enhancing both the task-specific performance of CLIP and its robustness to natural distribution shifts.
Additionally, we explore the few-shot adaptation of VLMs in a noisy setting, which better aligns with real-world scenarios where mislabeled samples may exist in the support set. We hope this work inspires future studies to focus on the generalization and robustness of pre-trained large-scale foundation models.

\end{document}